\pdfoutput=1

\documentclass[11pt]{article}

\usepackage[final]{acl}

\usepackage{times}
\usepackage{latexsym}

\usepackage[T1]{fontenc}

\usepackage[utf8]{inputenc}

\usepackage{microtype}

\usepackage{inconsolata}

\usepackage{graphicx}

\usepackage{amsmath}
\usepackage{algorithm,algorithmic}
\usepackage{multirow}
\usepackage{array}
\usepackage{subcaption}
\usepackage{graphicx}

\usepackage{forest}
\usepackage{tikz}
\usepackage{adjustbox}

\usepackage{bm}
\usepackage{bbm}
\usepackage{makecell}
\usepackage{wrapfig,lipsum}
\usepackage{caption}
\usepackage{tabularx}
\usepackage{multirow}
\usepackage{multicol}
\usepackage{wrapfig}
\usepackage{amssymb}
\usepackage{mathtools}

\usepackage{scalerel}
\newcommand\lightning{\scalerel*{\includegraphics{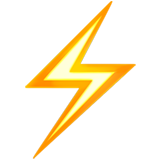}}{\rule{0pt}{1.2em}}}

\newcommand{\eg}{\emph{e.g.}}
\newcommand{\ie}{\emph{i.e.}}
\newcommand{\wrt}{\emph{w.r.t. }}

\newtheorem{remark}{Remark}

\title{\lightning Bypass Back-propagation: Optimization-based Structural Pruning for Large Language Models via Policy Gradient}

\author{
    \textbf{Yuan~Gao\textsuperscript{2}$^*$, ~Zujing~Liu\textsuperscript{1}$^*$, ~Weizhong~Zhang\textsuperscript{3}$^*$}, ~Bo~Du\textsuperscript{1}, ~Gui-Song~Xia\textsuperscript{2}$^\dag$\\
    \textsuperscript{1}School of Computer Science, Wuhan University \\
    \textsuperscript{2}School of Artificial Intelligence, Wuhan University \\
    \textsuperscript{3}School of Data Science, Fudan University \\
    \texttt{ethan.y.gao@gmail.com, weizhongzhang@fudan.edu.cn} \\
    \texttt{\{zujing.liu, dubo, guisong.xia\}@whu.edu.cn}
}

\begin{document}
\maketitle
{\let\thefootnote\relax\footnotetext{* Equal contribution.}
\let\thefootnote\relax\footnotetext{$\dagger$ Corresponding author.}}
\begin{abstract}
    Recent Large-Language Models (LLMs) pruning methods typically operate at the post-training phase without the expensive weight finetuning, however, their pruning criteria often rely on \textbf{heuristically hand-crafted metrics}, potentially leading to suboptimal performance. We instead propose a novel \textbf{optimization-based structural pruning} that learns the pruning masks in a probabilistic space directly by optimizing the loss of the pruned model. To preserve efficiency, our method \textbf{eliminates the back-propagation} through the LLM \emph{per se} during optimization, requiring only \textbf{the forward pass of the LLM}. We achieve this by learning an underlying \texttt{Bernoulli} distribution to sample binary pruning masks, where we decouple the \texttt{Bernoulli} parameters from LLM loss, facilitating efficient optimization via \emph{policy gradient estimator} without back-propagation. Thus, our method can 1) \emph{support global and heterogeneous pruning} (\ie, automatically determine different redundancy for different layers), and 2) \emph{optionally initialize with a metric-based method} (for our \texttt{Bernoulli} distributions). Extensive experiments conducted on LLaMA, LLaMA-2, LLaMA-3, Vicuna, and Mistral models using the C4 and WikiText2 datasets demonstrate the promising performance of our method in efficiency and effectiveness. Code is available at \url{https://github.com/ethanygao/backprop-free_LLM_pruning}.
    
\end{abstract}

\vspace{-2mm}
\section{Introduction}
\vspace{-1mm}

With the rapid development of Large Language Models \cite{brown2020language, achiam2023gpt} (LLMs) and their expanding across various applications, the efficiency of LLMs with vast parameters and complex architectures becomes crucial for practical deployment. In this paper, we aim to compress the LLM through structural pruning, which removes certain structural components such as channels and attention heads, \ie, \emph{Width Pruning} \cite{ma2023llm, muralidharan2024compact}, which is also our main concern, to reduce the model size with hardware-friendly acceleration.


Structural pruning methods in the pre-LLM era prune channels or layers via \emph{optimization}, using task loss back-propagation to determine pruning structures \cite{liu2018rethinking, blalock2020state}. These methods operate during training \cite{huang2018data, evci2020rigging} or post-training \cite{molchanov2019importance, wang2021neural}, where the latter is more efficient without weight updates. We focus on post-training pruning for efficiency.

However, the heavy computational and memory demands of LLMs make existing \emph{optimization-based pruning} methods less appropriate for efficiency. \emph{Metric-based pruning} is introduced to alleviate this issue, which directly prunes specific network components based on carefully designed criteria \cite{sun2023simple, das2023beyond}. Nonetheless, those criteria are often hand-crafted heuristically. As a result, metric-based pruning methods face challenges in achieving promising performance and generalizability, particularly at high pruning rates.

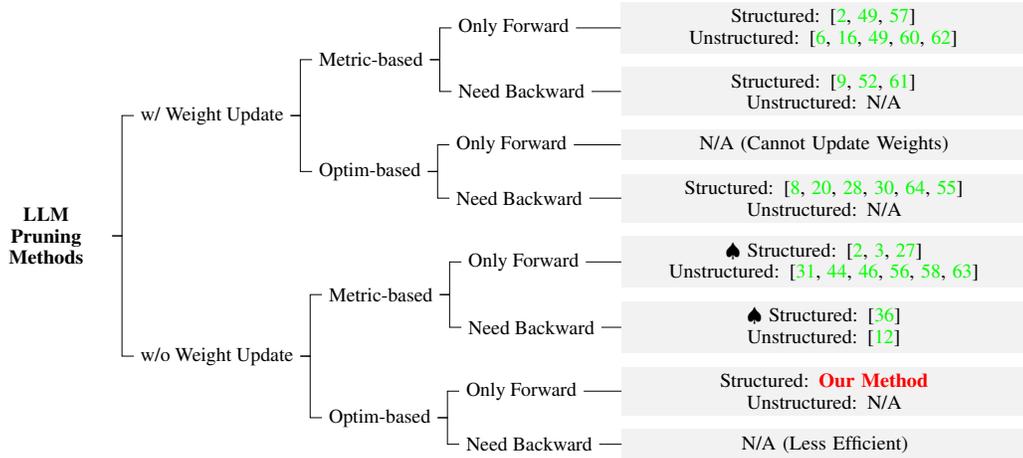
\begin{figure*}[t]
\vspace{-6mm}
\begin{adjustbox}{max width=\textwidth}
\small
\tikzset{
    basic/.style  = {draw, text width=3cm, align=center, font=\sffamily, rectangle},
    root/.style   = {align=center, text width=6em},
    tnode/.style = {thin, align=right, fill=gray!10, text width=0.45\textwidth, align=center, font=\footnotesize, },
    edge from parent/.style={draw=black, edge from parent fork right}
}

\begin{forest} 
for tree={
    edge path={
        \noexpand\path[\forestoption{edge}](!u.parent anchor) -- +(4pt,0) |- (.child anchor)\forestoption{edge label};},
    grow=0,
    reversed, 
    parent anchor=east,
    child anchor=west, 
    anchor=west,
}
[\textbf{LLM Pruning Methods}, root
   [ w/ Weight Update
        [Metric-based
            [Only Forward, l sep=8mm,
                [ \textbf{Struct}: \cite{an2024fluctuation, van2023llm, yang2024laco, shen2024search} \\ 
                \textbf{Unstruct}: \cite{bovza2024fast,  frantar2023sparsegpt, van2023llm, zeng2024multilingual, zhang2023efficient, meng2024alps, tan2024wrp}, tnode]
            ]
            [Need Backward, l sep=5.25mm,
                [ \textbf{Struct}: \cite{chen2024compressing, wei2024assessing, zhang2024loraprune} \\ 
                \textbf{Unstruct}: N/A, tnode]
            ]
        ]
        [Optim-based
            [Only Forward, l sep=8.4mm,
                [\textbf{Struct}: N/A \\ 
                \textbf{Unstruct}: \cite{bai2024sparsellm}, tnode]
            ]
            [Need Backward, l sep=5.7mm,
                [ \textbf{Struct}: \cite{chen2023lorashear, guo2023compresso, ko2023nash, li2024nuteprune, zhao2024apt, xia2023sheared, muralidharan2024compact} \\ 
                \textbf{Unstruct}: N/A, tnode]
            ]
        ]
   ]
   [w/o Weight Update
        [Metric-based
            [Only Forward, l sep=6.35mm,
                [$\spadesuit$ \textbf{Struct}: \cite{ashkboos2024slicegpt, kim2024shortened, dery2024everybody, song2024sleb} \\ 
                \textbf{Unstruct}: \cite{li2023sparse, shao2024one, sun2023simple, xu2024besa, yin2023outlier, zhang2023dynamic, lu2024alphapruning, li2024discovering}, tnode]
            ]
            [Need Backward, l sep=3.55mm,
                [$\spadesuit $ \textbf{Struct}: \cite{ma2023llm} \\ 
                \textbf{Unstruct}: \cite{das2023beyond}, tnode]
            ]
        ]
        [Optim-based
            [Only Forward, l sep=6.75mm,
                [ \textbf{Struct}: \textbf{\color{red} Our Method} \\ 
                \textbf{Unstruct}: N/A, tnode]
            ]
            [Need Backward, l sep=4mm,
                [ \textbf{Struct}: N/A \\ 
                \textbf{Unstruct}: \cite{fang2024maskllm}, tnode]
            ]
        ]
    ]
]
\end{forest}
\end{adjustbox}
\vspace{-6mm}
\caption{The taxonomy of our method among the LLM Pruning. Methods without weight update are used for comparison in our experiments (highlighted with $\spadesuit$), due to the constraints on time and memory efficiency, as well as the accessibility of large-scale finetuning datasets.}
\label{tab:taxonomy_detailed}
\vspace{-5.5mm}
\end{figure*}
Moreover, most \emph{metric-based pruning} methods typically prune the networks by manually-designed thresholds \cite{li2023sparse, zhang2023dynamic}. Although different layers of LLMs may have varying levels of redundancy \cite{yin2023outlier, xu2024besa}, \emph{achieving a global and heterogeneous pruning strategy is challenging with metric-based approaches}. This is due to the significantly varying magnitudes of the manually designed metrics across layers, making it laborious or even impossible to set proper pruning threshold for each layer\footnote{As a practical compromise, most metric-based methods conduct a homogeneous/uniform pruning rate for all the layers, which violates the fact that different layers could possess the different amount of redundancy.}. 

The above analysis leads to a natural question: \emph{Can we attain the performance of \textbf{optimization-based methods} that facilitate global and heterogeneous pruning without relying on hand-crafted heuristics, while preserving a similar cost with the \textbf{metrics-based methods} that is affordable on a single commercial GPU?}
\vspace{-0.8mm}

In view of the above analysis, our proposed method is essentially a novel lightweight \emph{optimization-based method}, where it 1) efficiently avoids the back-propagation through the heavy LLM, 2) optionally can be initialized by an arbitrary metric-based approach. Particularly, our pruning efficiency is ensured via a \emph{policy gradient estimator} \cite{williams1992simple}, requiring only the LLM \textbf{forward pass} without back-propagation, which is analogous to many efficient metric-based methods and requires the same memory overhead, such as \cite{sun2023simple, an2024fluctuation}. Moreover, our method unifies the pruning of the entire LLM into a probabilistic space (optionally initialized by an arbitrary metric-based approach), eliminating the magnitude difference issue of most metric-based methods and therefore directly facilitating global and heterogeneous pruning across the entire LLM.

Specifically, we formulate our pruning as a binary mask optimization problem \cite{srinivas2017training}, where the binary masks determine whether to prune the corresponding structures via element-wise product. To efficiently learn those binary masks, we construct an underlying probabilistic space of \texttt{Bernoulli} distributions to sample them. By decoupling the \texttt{Bernoulli} parameters from sampled masks, our method disentangles these parameters from the LLM loss, enabling efficient optimization via \emph{policy gradient estimator}, bypassing back-propagation\footnote{Note that our formulation can also be interpreted from a reinforcement learning (with dense rewards) perspective in terms of Markov Decision Process, detailed in Appendix \ref{app:rl}}. Moreover, the probabilistic modeling of \texttt{Bernoulli} distribution facilitates global and heterogeneous pruning across the LLM. 

The taxonomy of our methods is illustrated in Fig. \ref{tab:taxonomy_detailed}. In the experiments, our method is compared with SOTA structural pruning methods that \emph{do not update the model weight simultaneously}, due to the constraints on \textbf{time and memory efficiency}\footnote{After pruning, it is affordable to finetune the pruned smaller model on a single commercial GPU. The performance with pruning then finetuning is included in our experiments.}. We extensively validate our methods using the C4 \cite{raffel2020exploring} and WikiText2 \cite{merity2016pointer} datasets on popular LLaMA \cite{touvron2023llama1}, LLaMA-2 \cite{touvron2023llama2}, LLaMA-3 \cite{dubey2024llama}, Vicuna \cite{chiang2023vicuna}, and Mistral \cite{jiang2023mistral} models with various parameter sizes, pruning rates, and initializations, showing the promising performance and efficiency. For example, our method outperforms the SOTA methods regarding both perplexity and zero-shot performance and operates only 2.7 hours with about 35GB memory on a single A100 GPU to prune the LLaMA-2-13B model. Our method exhibits the following features:
\begin{itemize}
\vspace{-3mm}
\item \textbf{Accuracy}, ensured by 1) our \emph{optimization-based pruning} without heuristically hand-crafted metrics, which optionally takes metric-based pruning as initialization for a better convergence, and 2) the \emph{global and heterogeneous pruning}, as supported by our probabilistic modeling of the pruning masks. 
\vspace{-3mm}
\item \textbf{Efficiency} (regarding both computations and memory), achieved by the \emph{policy gradient estimator} for back-propagation-free and forward-only optimization \wrt the heavy LLMs.
\vspace{-3mm}
\end{itemize}

\vspace{-1.4mm}
\section{Related Work}
\vspace{-1.6mm}

Pruning has proven effective in traditional deep neural networks \cite{han2015deep, frankle2018lottery, kurtic2022optimal, liu2019metapruning, he2018amc}, and extensive research has been conducted on this topic. Typically, post-pruning performance is restored or even enhanced through full-parameter fine-tuning \cite{liu2018rethinking, blalock2020state}. However, for large language models (LLMs) with vast parameters, full-parameter fine-tuning is computationally expensive and often impractical. To overcome this challenge, various pruning strategies \cite{ma2023llm, zhang2024loraprune, sun2023simple, ashkboos2024slicegpt, frantar2023sparsegpt} have been developed for LLMs in recent years. These strategies can be categorized into metric-based pruning and optimization-based pruning.

\textbf{Metric-based Pruning.} Metric-based pruning methods focus on designing importance metrics for model weights or modules. \cite{sun2023simple} introduces a pruning metric by considering both the magnitude of weights and activations. LLM-Pruner \cite{ma2023llm} eliminates coupled structures with low weight importance via loss change. These methods use pre-defined pruning metrics and often face challenges with high pruning rates. 
\cite{dery2024everybody} proposed a structured pruning method using only forward passes with promising performance. It regresses the heuristically hand-crafted criteria, \eg, the utility of the pruned sub-networks, and makes assumptions that may not hold universally, \eg, the network's utility as a linear sum of building elements' utilities, and their utility being consistent/average-able across sub-networks.

Metric-based pruning methods use predefined criteria, potentially leading to suboptimal performance. Our optimization-based pruning framework, inspired by Neural Architecture Search (NAS) \cite{liu2018darts}, directly optimizes the loss function to identify the optimal pruned architectures while achieving higher efficiency through policy gradient optimization compared to conventional NAS that rely on back-propagation.

\textbf{Optimization-based Pruning.} Optimization-based pruning methods focus on determining the model mask in an optimized manner and also involve model weight updating. Sheared LLaMA \cite{xia2023sheared} learns pruning masks to find a subnetwork that fits a target architecture with full-parameters updating. \cite{guo2023compresso, chen2023lorashear, zhao2024apt} utilize LoRA \cite{hu2021lora} in the pruning process with weight updating. 

However, these methods rely on costly back-propagation for optimization and weight updating. Instead, we propose using policy gradient estimation in the optimization process as an alternative, significantly reducing the computational demands.

\vspace{-2mm}
\section{Methodology}
\vspace{-2mm}


\begin{figure*}
\vspace{-7mm}
    \centering
    \includegraphics[width=0.95\textwidth]{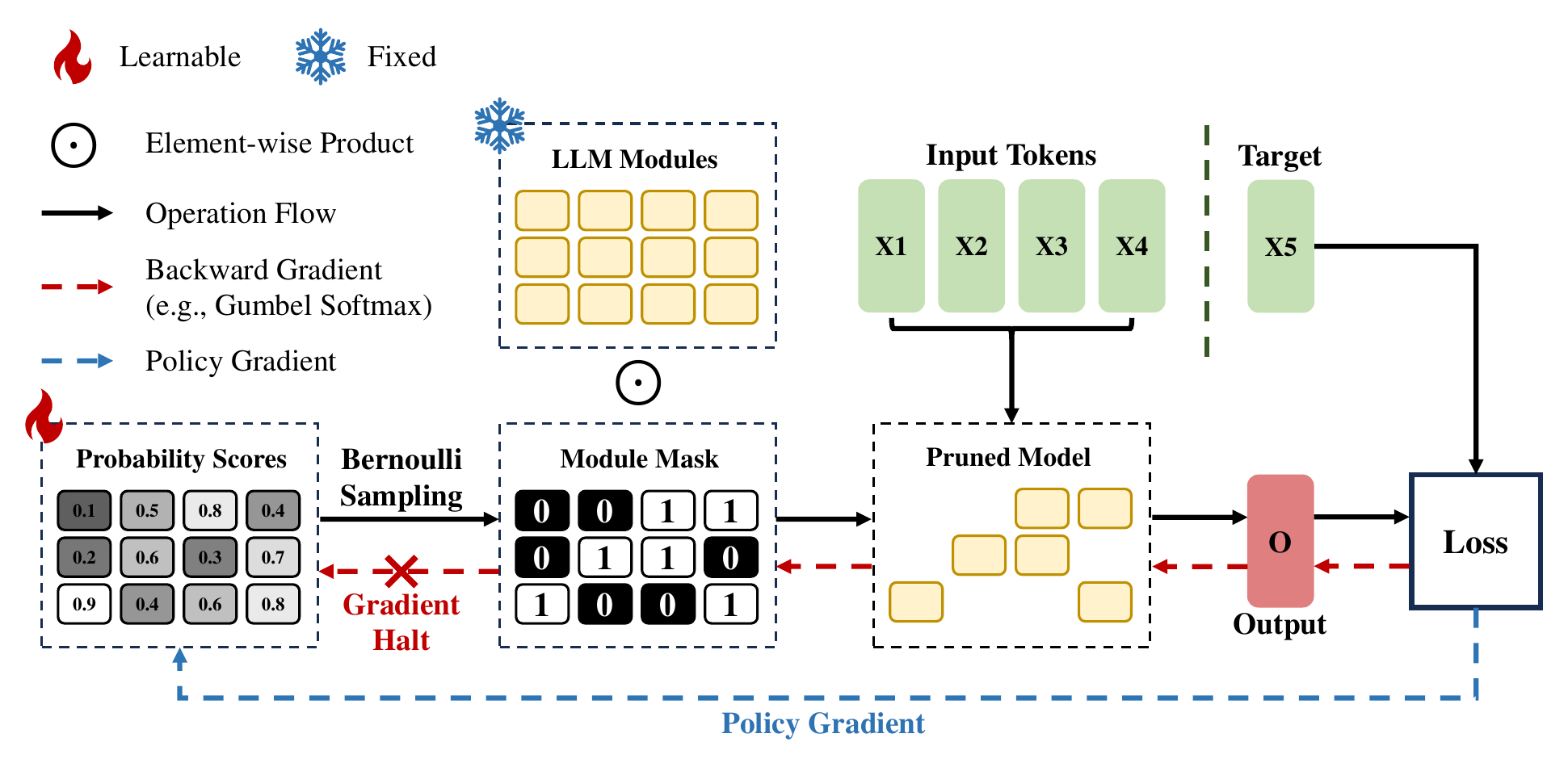}
    \vspace{-5mm}
    \caption{The overview of our method. We formulate LLM pruning as optimizing underlying \texttt{Bernoulli} distributions that sample binary masks. Being different from the conventional back-propagation method (\eg, through \emph{Gumbel Softmax} as shown by the {\color{red} red-dashed-arrows}), our formulation decouples the masks and the \texttt{Bernoulli} parameters from the LLM loss (see Eq. \eqref{eq:Phi} and Remark \ref{remark3}), facilitating efficient and unbiased \emph{policy gradient} (the {\color{blue} blue-dashed-arrow}) without back-propagation through the LLM (see Eq. \eqref{eq3:pge} and Remark \ref{remark4}).} 
    \label{fig:method-overview}
\vspace{-5.5mm}
\end{figure*}

We introduce our optimization-based pruning for LLMs, which is efficient without back-propagation through the LLM, illustrated in Fig. \ref{fig:method-overview}.

\vspace{-2mm}
\subsection{Pruning via Probabilistic Mask Modeling} \label{sec:mask_modeling}
\vspace{-1mm}

We formulate the network pruning as seeking binary masks \cite{srinivas2017training} to determine whether the corresponding structure should be pruned or not. Those binary masks are further modeled by/sampled from the \texttt{Bernoulli} distributions stochastically. Such formulation possesses several merits: 1) the probabilistic \texttt{Bernoulli} modeling facilitates global and heterogeneous pruning across the entire LLM; 2) our stochastical sampling decouples \texttt{Bernoulli} parameters and the sampled masks from LLM loss empowering an efficient \emph{policy gradient} optimization without back-propagate through the LLM (see Sect. \ref{sec:pge}); and 3) the mask formulation enables flexible pruning at channels, heads (of Multi-Head Attention, MHA), and layers.

We denote the calibration dataset with $N$ \emph{i.i.d.} samples as $\mathcal{D} \!=\! \{ (\mathbf{x}_i, \mathbf{y}_i)\}^N_{i=1} $, $\mathbf{w} \!=\!  \{ \mathbf{w}_i\}^n_{i=1}$ as the complete and non-overlapped modules of a LLM with model size $n$, and $\mathbf{m} \!=\! \{ \mathbf{m}_i\}^n_{i=1} \!\in\! \{0,1\}^n$ as the corresponding binary masks, where $\mathbf{m}_i \!=\! 0$ implies $\mathbf{w}_i$ is pruned and otherwise retained. Note that $\mathbf{w}_i$ and $\mathbf{m}_i$ can be defined at various granularities such as channels, heads, and layers\footnote{For the channel and head granularity, we prune the dimensions of the hidden states following \cite{ma2023llm} while preserving output channels of each block to maintain residue connections(see Appendix \ref{app:imp_details}).}. Then, our structural pruning of LLMs can be formulated as a binary optimization with constraints:
{\setlength\abovedisplayskip{3pt}
\setlength\belowdisplayskip{3pt}
\begin{align}
\min_{\mathbf{m}}\mathcal{L}( \mathcal{D};&\mathbf{w} \!\odot\!  \mathbf{m} ) \!:=\! \frac{1}{N} \! \sum_{i=1}^N \ell ( f(\mathbf{x}_i; \mathbf{w} \!\odot\!  \mathbf{m} ), \mathbf{y}_i), \nonumber \\
\text{s.t. } &\left \| \mathbf{m}  \right \|_1 \leq rn \ \text{ and } \  \mathbf{m} \in \{ 0,1\}^n.
\label{eq1-binary_optim}
\end{align}}

where $f(\cdot ; \mathbf{w} \odot \mathbf{m})$ is the pruned network, $\ell (\cdot, \cdot)$ is the loss function, \eg, the cross-entropy loss, and $r$ is the target pruning rate. We note that the binary optimization problem Eq. \eqref{eq1-binary_optim}, \ie, finding optimal masks $\mathbf{m}$ from the discrete and exponentially growing solution space, is typically NP-hard.

Therefore, we relax the discrete optimization using a probabilistic approach, by treating $n$ masks as binary \emph{random variables} sampled from $n$ underlying \texttt{Bernoulli} distributions with parameters $\mathbf{s} = \{ \mathbf{s}_i\}^n_{i=1} \in [0, 1]^n$. This yields the conditional distribution of $\textbf{m}$ over $\mathbf{s}$:
\begin{equation}\label{eq2-mask_prob}
\setlength\abovedisplayskip{3pt}
\setlength\belowdisplayskip{3pt}
p(\mathbf{m}|\mathbf{s})=\prod_{i=1}^n(s_i)^{m_i}(1-s_i)^{1-m_i}.
\end{equation}
By relaxing the $\ell_1$ norm in Eq. \eqref{eq1-binary_optim} by its expectation, \ie, $\left \| \mathbf{m}  \right \|_1 \!\approx\! \mathbb{E}_{\mathbf{m} \sim p(\mathbf{m}|\mathbf{s})}\! \left \| \mathbf{m} \right \|_1 \!\!=\! \sum_{i=1}^{n} \! s_i \!=\! \mathbf{1}^{\top}\!\mathbf{s}$, we have the following expected loss minimization problem:
\begin{equation} \label{eq2-relaxed_problem}
\setlength\abovedisplayskip{3pt}
\setlength\belowdisplayskip{3pt}
\begin{aligned}
    \min_{\mathbf{s}} \  &\mathbb{E}_{p(\mathbf{m}|\mathbf{s})} \mathcal{L} (\mathcal{D};\mathbf{w} \!\odot\! \mathbf{m}), \\
\text{s.t.} \ \mathbf{1}^{\top}&\mathbf{s} \leq rn \  \text{ and } \ \mathbf{s} \in [0,1]^n.
\end{aligned}
\end{equation}

\begin{remark}
Problem \eqref{eq2-relaxed_problem} is a continuous relaxation of the discrete Problem \eqref{eq1-binary_optim}. The feasible region of \eqref{eq2-relaxed_problem} is the intersection of the cube $[0,1]^n$ and the half-space $\mathbf{1}^{\top}\mathbf{s} \leq rn$. Moreover, the parameterization of \eqref{eq2-relaxed_problem} in the probabilistic space facilitates automatically learning the redundancy across different layers for global and heterogeneous pruning.
\end{remark}
\vspace{-3mm}

\vspace{-1mm}
\subsection{Policy Gradient Optimization} \label{sec:pge}
\vspace{-1mm}

Conventional neural network training paradigm usually adopts back-propagation to estimate the gradient of Eq. \eqref{eq2-relaxed_problem}, \eg, through Gumbel Softmax \cite{maddison2016concrete,dupont2022extracting} which reparameterizes the mask $\mathbf{m}$ as a function of $\mathbf{s}$, \ie, $m_i \!=\! \phi(s_i)$ or  $m_i \!=\! \phi(s_i, \epsilon)$ with $\epsilon \!\sim\! \mathcal{N}(0,1)$. However, the back-propagation has the following intrinsic issues in LLM pruning. 

\vspace{-1.5mm}
\begin{remark}
\textbf{Intrinsic issues of back-propagation in LLM pruning}: 1) the back-propagation is computationally expensive and memory-intensive; 2) the computation of gradients can not be satisfied by using the sparsity in $\mathbf{m}$, \ie, $\frac{\partial m_i}{\partial s_i} \neq 0$ even if $m_i = 0$. In other words, one has to go through the full model for back-propagation even when lots of the LLM modules have been masked.
\label{remark:backprop}
\end{remark}
\vspace{-1.6mm}

Now we present our efficient (back-propagation-free) and unbiased optimization for Problem \eqref{eq2-relaxed_problem}. We propose using Policy Gradient Estimator (PGE) for the gradient estimation with only forward pass, avoiding the pathology of the chain-rule estimator. Specifically, in order to update the \texttt{Bernoulli} parameters $\mathbf{s}$, we have the objective $\Phi(\mathbf{s})$:

\vspace{-3mm}
\begin{equation} 
\setlength\abovedisplayskip{1pt}
\begin{aligned}
\Phi(\mathbf{s}) = \mathbb{E}_{p(\mathbf{m}|\mathbf{s})} \mathcal{L} (\mathbf{m}) & \!=\! \!\int\! p(\mathbf{m}|\mathbf{s}) \mathcal{L}(\mathbf{m}) \text{d}\mathbf{m}, \\
\text{s.t.} \ \mathbf{1}^{\top}\mathbf{s} \leq rn  & \text{ and } \ \mathbf{s} \in [0,1]^n.
\end{aligned}\label{eq:Phi}
\vspace{-3mm}
\end{equation}

Our key idea is that in Eq. \eqref{eq:Phi}, the score vector $\mathbf{s}$ only appears in the conditional probability $p(\mathbf{m}|\mathbf{s})$ for sampling $\mathbf{m}$, which is decoupled from the network loss term $\mathcal{L}(\mathbf{m})$, short for $\mathcal{L}(\mathcal{D};\mathbf{w} \!\odot\! \mathbf{m})$.

\vspace{-1mm}
\begin{remark} \label{remark3}
\textbf{Differences with Gumbel Softmax}: 1) As shown in Eq. \eqref{eq:Phi}, our PGE formulates the mask $\mathbf{m}$ as a random variable which is only related to the distribution $\mathbf{s}$ through the conditional probability $p(\mathbf{m}|\mathbf{s})$ of probabilistic sampling. Thus, the expensive back-propagation through the LLM can be omitted in gradient estimation using the PGE. In contrast, for the Gumbel Softmax estimator, $\mathbf{m}$ is a function of $\mathbf{s}$, requiring the back-propagation through the whole network (see the {\color{blue} blue} and {\color{red} red} gradient flows in Fig. \ref{fig:method-overview}). 2) As a result, Gumbel Softmax is challenged by the back-propagation issues discussed in Remark \ref{remark:backprop}. 3) Gumbel Softmax is known to be \emph{biased} especially when the temperature is high \cite{huijben2022review}. 4) The vanilla PGE might suffer from \emph{large variance} \cite{liu2020improved}, so we exploit a variance-reduced PGE discussed later in Eq. \eqref{eq:update_s} with theoretical analysis and empirical ablations in Appendices \ref{app:vrpg_theorical} and \ref{app:vrpg_ablation}.
\end{remark}
\vspace{-1mm}

Specifically, the optimization of Eq. \eqref{eq:Phi} via the policy gradient estimator holds that:
{\setlength\abovedisplayskip{3pt}
\setlength\belowdisplayskip{3pt}
\begin{align} 
\nabla_{\mathbf{s}} \Phi(\mathbf{s}) & \!=\! \!\!\int\!\! \mathcal{L}(\mathbf{m})\!\nabla_{\mathbf{s}}p(\mathbf{m}|\mathbf{s})\!+\!\underbrace{p(\mathbf{m}|\mathbf{s})\!\nabla_{\mathbf{s}}\mathcal{L}(\mathbf{m})}_{=\ 0} \text{d}\mathbf{m}  \nonumber \\
& \!=\! \!\!\int\!\! \mathcal{L}(\mathbf{m}) p(\mathbf{m}|\mathbf{s})\!\nabla_{\mathbf{s}}\log(p(\mathbf{m}|\mathbf{s})) \text{d}\mathbf{m} \nonumber \\
& \!=\! \mathbb{E}_{p(\mathbf{m}|\mathbf{s})} \mathcal{L}(\mathbf{m}) \!\nabla_{\mathbf{s}}\log(p(\mathbf{m}|\mathbf{s})).
\label{eq3:pge}
\end{align}}


The final equality provides conclusive proof that $\mathcal{L}(\mathbf{m}) \nabla_{\mathbf{s}}\log(p(\mathbf{m}|\mathbf{s}))$ is an unbiased stochastic gradient for $\Phi(\mathbf{s})$. 

\vspace{-1.5mm}
\begin{remark} \label{remark4}
\textbf{The efficiency of Eq. \eqref{eq3:pge}}: 1) Equation \eqref{eq3:pge} can be computed purely with forward propagation. 2) The computation cost for the gradients, \ie, $\nabla_{\mathbf{s}}\log(p(\mathbf{m}|\mathbf{s})) = \frac{\mathbf{m}-\mathbf{s}}{\mathbf{s}(1-\mathbf{s})}$, is negligible. Therefore, our PGE is much efficient compared to the backward-propagation-based estimators. 
\end{remark}
\vspace{-1.5mm}

The stochastic gradient descent algorithm is:
\vspace{-1mm}
\begin{equation} \label{eq4-sgd}
\setlength\abovedisplayskip{3pt}
\setlength\belowdisplayskip{3pt}
\begin{aligned}
& \mathbf{s} \leftarrow \textbf{proj}_{\mathcal{C}}(\mathbf{z}), \ \\
\mathbf{z} := \mathbf{s} - \eta \mathcal{L}( & \mathcal{D}_B;\mathbf{w} \!\odot\! \mathbf{m}) \nabla_{\mathbf{s}}\log(p(\mathbf{m}|\mathbf{s})).
\end{aligned}
\end{equation}
where $\mathcal{D}_B \!=\!\{ (\mathbf{x}_i, \mathbf{y}_i)\}^B_{i=1}$ is batch samples from $\mathcal{D}$ with batch size $B$, and $\mathcal{L}(\mathcal{D}_B;\mathbf{w} \!\odot\! \mathbf{m})$ is the loss on $\mathcal{D}_B$ with the pruned model by masks $\mathbf{m}$. The projection operator $\textbf{proj}_{\mathcal{C}}(\cdot)$ is to ensure the updated scores $\mathbf{s}$ to be constrained in the feasible domain $\mathcal{C} = \left \{ \mathbf{1}^{\top}\mathbf{s} \leq K \right\}  \bigcap \left \{\mathbf{s} \in [0,1]^n \right \}$. We implement the projection operator from \cite{wang2013projection}, the details of which can be found in Appendix \ref{app:proj_op}.

Policy gradient might suffer from large variance \cite{liu2020improved}. To reduce the variance for fast and stable training, we minus a moving average baseline \cite{zhao2011analysis} which is calculated by 1) obtaining the averaged loss of multiple sampling trials, then 2) taking the moving average of the current and the previous losses given a window size. Denote the baseline as $\delta$, given window size $T$ (set to 5), and mask sampling times $N_s$ (set to 2), we update $\mathbf{s}$ in each training step via Eqs. \eqref{eq:update_s} and \eqref{eq:update_baseline}. The theoretical analysis and empirical ablations can be found in Appendices \ref{app:vrpg_theorical} and \ref{app:vrpg_ablation}.
\setlength\abovedisplayskip{3pt}
\setlength\belowdisplayskip{3pt}
\begin{align}
\mathbf{s} & \leftarrow \textbf{proj}_{\mathcal{C}}(\mathbf{z}) \ \text{with} \ \mathbf{z} := \mathbf{s} - \eta \bigg[\frac{1}{N_s} \label{eq:update_s} \\
& \sum_{i=1}^{N_s} \big( \mathcal{L}(\mathcal{D}_B; \mathbf{w} \!\odot\! \mathbf{m}^{(i)}) \notag {\color{red} - \delta} \big) \nabla_{\mathbf{s}}\log(p(\mathbf{m}^{(i)}|\mathbf{s})) \bigg]. \\
\delta & \leftarrow \frac{T-1}{T}\delta + \frac{1}{N_sT} \sum_{i=1}^{N_s} \mathcal{L}(\mathcal{D}_B;\mathbf{w} \!\odot\! \mathbf{m}^{(i)}). \label{eq:update_baseline}
\end{align}


Our efficient pruning algorithm is summarized in Appendix \ref{app:proj_op}. Note that our formulation can be interpreted as a dense rewards reinforcement learning problem, detailed in Appendix \ref{app:rl}.

\textbf{Initialization.} Algorithms based on policy gradient usually require an effective initialization to get enhanced results. In this context, previous hand-crafted pruning metric can be applied to initialize the probability of each module: $\mathbf{s}_0 \!\leftarrow\! \sigma(\mathbf{x})$, in which $\mathbf{x}$ can be any pruning metric derived from existing method, $\mathbf{s}_0$ represents the initial probability assigned to each module, and $\sigma$ symbolizes a non-linear transformation. \textbf{We note that initializing from a prior metric-based method is only an option, while a random initialization strategy can already produce good performance.} Please refer to different initializations $\mathbf{x}$ and transformations $\sigma$ discussed in Appendices \ref{app:diff_init} and \ref{app:proj_strategy}.


    
\textbf{Applicability of PGE in Learning Pruning Masks.} We note that the precision of PGE may not match that of conventional back-propagation. Given that we are learning the \textbf{binary} masks $\mathbf{m}$ (distinct from the \textbf{float} weights), it is expected that the precision requirement of $\mathbf{s}$ can be modest. Moreover, our PGE is unbiased (compared to the biased Gumbel Softmax). These factors make the PGE suitable for learning the masks, which is empirically validated with extensive experiments. We also compared the results of using PGE and Gumbel Softmax respectively in Sect. \ref{sec:gumbel}.

\textbf{Practical applicability of our efficient pruning method.} Our method is particularly effective in memory-constrained settings where GPU memory may only accommodate inference for the unpruned model. This addresses a growing practical challenge, as state-of-the-art models continue to expand while many researchers and institutions face hardware limitations. Moreover, the pruned smaller model remains affordable to fine-tune under these limitations. We discuss these practical implications in detail in Appendix \ref{app:pratical}.

\begin{table*}[t]
\vspace{-5mm}
\centering
\small
\begin{tabular}{@{}l||c||c|c||c|c||c||c|c@{}}
\hline
\hline
    \multirow{2}{*}{Method} & \multirow{2}{*}{PruneRate} & \multicolumn{2}{c||}{LLaMA} & \multicolumn{2}{c||}{LLaMA-2} & LLaMA-3 & \multicolumn{2}{c}{Vicuna} \\
    \cline{3-9}
    & & 7B & 13B & 7B & 13B & 8B & 7B & 13B \\
    \hline
    
    Dense  & 0$\%$ & 12.62 & 10.81 & 12.19 & 10.98 & 14.14 & 16.24 & 13.50 \\
    \hline

    LLM-Pruner & \multirow{4}{*}{30$\%$} & 38.41 & 24.56 & 38.94 & 25.54 & 40.18 & 48.46 & 31.29 \\
    SliceGPT & & - & - & 40.40 & 30.38 & 183.94 & 52.23 & 57.75 \\
    Bonsai & & 30.49 & 26.24 & 39.01 & 24.23 & 80.89 & 44.28 & 54.16 \\
    Wanda-sp & & 98.24 & 25.62 & 49.13 & 41.57 & 92.14 & 57.60 & 80.74 \\
    Ours & & \textbf{25.61} & \textbf{19.70} & \textbf{28.18} & \textbf{21.99} & \textbf{38.99} & \textbf{34.51} & \textbf{26.42} \\
    \hline


    LLM-Pruner & \multirow{4}{*}{40$\%$} & 72.61 & 36.22 & 68.48 & 37.89 & 70.60 & 88.96 & 46.88 \\
    SliceGPT & & - & - & 73.76 & 52.31 & 353.09 & 89.79 & 130.86 \\
    Bonsai & & 60.65 & 58.17 & 69.18 & 50.97 & 204.61 & 95.32 & 272.10 \\
    Wanda-sp & & 110.10 & 165.43 & 78.45 & 162.50 & 213.47 & 85.51 & 264.22 \\
    Ours & & \textbf{42.96} & \textbf{28.12} & \textbf{39.81} & \textbf{31.52} & \textbf{63.85} & \textbf{51.86} & \textbf{43.59} \\
    \hline


    LLM-Pruner & \multirow{4}{*}{50$\%$} & 147.83 & 67.94 & 190.56 & 72.89 & 145.66 & 195.85 & 91.07 \\
    SliceGPT & & - & - & 136.33 & 87.27 & 841.20 & 160.04 & 279.33 \\
    Bonsai & & 275.63 & 148.92 & 216.85 & 146.38 & 440.86 & 180.75 & 424.33 \\
    Wanda-sp & & 446.91 & 406.60 & 206.94 & 183.75 & 413.86 & 242.41 & 373.95 \\
    Ours & & \textbf{72.02} & \textbf{49.08} & \textbf{65.21} & \textbf{52.23} & \textbf{119.75} & \textbf{71.18} & \textbf{68.13} \\
\hline
\hline
\end{tabular}
\vspace{-3mm}
\caption{Results (perplexity) on \emph{channels and heads} pruning. Our method is initialized by Wanda-sp (please also refer to Sect. \ref{sec:init} and Appendix \ref{app:diff_init} for a detailed discussion about initializations). All the methods are calibrated using the C4 dataset and validated on the WikiText2 dataset \wrt perplexity.}
\label{tab:channel_ppl_results}
\vspace{-3mm}
\end{table*}

\vspace{-2mm}
\section{Experiments}
\vspace{-2mm}

We conduct extensive experiments to validate the promising performance of the proposed method, across \textbf{different LLM models with various sizes}, \textbf{pruning rates}, and \textbf{initializations} (in the ablation analysis). First, we detail our experimental setups in Sect. \ref{sec:setup}. After that, our main results against the state-of-the-art methods for channels and heads pruning are shown in Sect. \ref{sec:HCPruning}. We illustrate the zero-shot performance in Sect. \ref{sec:zero-shot}, Appendices \ref{app:finetune} and \ref{app:lamma2}. Our method runs 2.7 hours for LLaMA-2-13B with a similar GPU memory (\ie, $\sim$35GB) as Wanda-sp \cite{an2024fluctuation} as shown in Appendix \ref{app:time_memory}. Considering the constraints on computations and memory, we compare with the state-of-the-art methods without in-pruning weight update, and report the \emph{pruning then finetuning} performance in Appendix \ref{app:finetune}, as it becomes affordable to finetune a smaller pruned model. We also show generated samples of the pruned models in Table \ref{tab:gen_samples} of Appendix \ref{app:gen_samples} and multiple-run statistics of our method in Appendix \ref{app:error_bar}.

\begin{table*}[t]
\centering
\small
\begin{tabular}{@{}l||c||c||c|c|c|c|c||c@{}}
\hline
\hline
Method & PruneRate & PPL $\downarrow$ & PIQA & HellaSwag & WinoGrande & ARC-e & ARC-c & Average \\
\hline

Dense & 0$\%$ & 14.14 & 79.71 & 60.19 & 72.61 & 80.09 & 50.34 & 68.59 \\
\hline

LLM-Pruner & \multirow{5}{*}{30$\%$} & 40.18 & 71.38 & 37.84 & 55.64 & 57.78 & 27.21 & 49.97 \\
SliceGPT &                           & 183.94 & 68.34 & \textbf{53.92} & 57.22 & 49.41 & 28.07 & 51.39 \\
Bonsai &                             & 80.89 & 64.53 & 36.10 & 55.09 & 47.64 & 22.52 & 45.18 \\
Wanda-sp &                           & 92.14 & 59.74 & 31.46 & 52.64 & 44.02 & 19.88 & 41.55 \\
Ours &                               & \textbf{38.99} & \textbf{72.25} & 43.56 & \textbf{59.04} & \textbf{59.85} & \textbf{29.44} & \textbf{52.83} \\
\hline

LLM-Pruner & \multirow{5}{*}{40$\%$} & 70.60 & 66.26 & 31.90 & 54.06 & 49.74 & 22.52 & 44.90 \\
SliceGPT &                           & 353.09 & 61.53 & \textbf{39.98} & 52.80 & 36.66 & \textbf{25.17} & 43.23 \\
Bonsai &                             & 204.61 & 58.81 & 29.43 & 48.93 & 33.21 & 18.15 & 37.71 \\
Wanda-sp &                           & 213.47 & 56.58 & 27.46 & 50.35 & 32.07 & 17.06 & 36.70 \\
Ours &                               & \textbf{63.85} & \textbf{67.63} & 37.36 & \textbf{56.91} & \textbf{50.67} & 24.91 & \textbf{47.50} \\
\hline

LLM-Pruner & \multirow{5}{*}{50$\%$} & 145.65 & 61.15 & 29.10 & \textbf{51.93} & 39.98 & 19.36 & 40.30 \\
SliceGPT &                           & 841.20 & 56.37 & \textbf{32.66} & 48.38 & 32.45 & \textbf{22.10} & 38.39 \\
Bonsai &                             & 440.86 & 55.66 & 26.94 & 50.51 & 30.64 & 17.83 & 36.32 \\
Wanda-sp &                           & 413.86 & 55.39 & 27.07 & 49.72 & 29.59 & 18.26 & 36.01 \\
Ours &                               & \textbf{119.75} & \textbf{62.51} & 30.89 & 51.85 & \textbf{41.12} & 20.65 & \textbf{41.40} \\

\hline
\hline
\end{tabular}
\vspace{-3mm}
\caption{Perplexity (PPL) and zero-shot accuracies ($\%$) of LLaMA-3-8B for 5 zero-shot tasks.}
\label{tab:zero_shot_llama3}
\vspace{-5mm}
\end{table*}

\vspace{-2mm}
\subsection{Experimental Setups} \label{sec:setup}
\vspace{-1mm}

\textbf{Structural Granularities for Pruning.} We validate our method on \emph{Head and Channel Granularity} for pruning, \ie, Width Pruning. 
For the effects of different initializations, we extensively investigated them in Sect. \ref{chap:ablation} and Appendices \ref{app:diff_init} and \ref{app:proj_strategy}.

\emph{Head and Channel Granularity}: We follow \cite{ma2023llm, an2024fluctuation} to prune the \emph{heads} of the multi-head attention (MHA) modules and the \emph{channels} of the MLP modules in Sect. \ref{sec:HCPruning}. 
We initialize our methods with an efficient metric-based structural pruning method, \ie, Wanda-sp \cite{an2024fluctuation}. Our method is compared to the state-of-the-art Wanda-sp \cite{an2024fluctuation}, LLM-Pruner \cite{ma2023llm}, SliceGPT \cite{frantar2023sparsegpt}, and Bosai \cite{dery2024everybody}.

Additionally, we also validate our method on \emph{Layer Granularity}, \ie, Depth Pruning \cite{kim2024shortened, song2024sleb}, by pruning the entire transformer layer, shown in Appendix \ref{app:LPruning}.


\textbf{LLM Models and Sizes.} LLaMA-\{7B, 13B\} \cite{touvron2023llama1}, LLaMA-2-\{7B, 13B\} \cite{touvron2023llama2}, LLaMA-3-8B \cite{dubey2024llama}, Vicuna-\{7B, 13B\} \cite{chiang2023vicuna}, and Mistral-7B-Instruct-v0.3 \cite{jiang2023mistral} are used as the source models in our experiments.

\textbf{Pruning Rate.} Promising performance with a high pruning rate could be challenging to obtain when employing metric-based pruning, owing to the heuristically designed metrics. To validate the superior performance of our optimization-based pruning under this situation, we select high pruning rates ranging from 30\% to 50\%, \ie, \textbf{structurally} removing 30\% to 50\% model parameters. 

\textbf{Datasets.} We perform the experiments following the cross-dataset settings in \cite{sun2023simple}, where the C4 dataset \cite{raffel2020exploring} is used for training and the WikiText2 dataset \cite{merity2016pointer} is used for evaluation. This challenging cross-dataset setup potentially better reflects the generalization of the pruned model.

\textbf{Training and Evaluation Details.} We update the underlying \texttt{Bernoulli} distributions (for mask sampling) simply using SGD with a learning rate of 6e-3 for LLaMA-3 experiments and 2e-3 for the remaining. The batch size is fixed to 8 and we train our lightweight policy gradient estimator for 1 epoch on the C4 dataset with 120K segments, where each segment has a sequence length of 128. Ablations on calibration dataset size are also conducted, detailed in Appendix \ref{app:calib_data_size}.

To reduce the evaluation variance, we deterministically generate the pruned evaluation architecture, \ie, given a pruning rate $r$, we first rank all the $\mathbf{s}$, then deterministically set $\mathbf{m}$ corresponding to the minimal $r$ of $\mathbf{s}$ as 0 (otherwise 1). We report the perplexity on the WikiText2 dataset using a sequence length of 128. Given a tokenized sequence $\mathbf{X} = (x_0, x_1, \dots, x_t)$, the perplexity of $\mathbf{X}$ is:
\begin{align}
    \text{Perplexity}(\mathbf{X}) = \text{exp} \left \{ - \frac{1}{t} \sum_{i}^t \log p_{\theta} (x_i|x_{<i}) \right \}, \nonumber
\end{align}
where $\log p_{\theta} (x_i|x_{<i})$ is the log-likelihood of token $x_i$ conditioned on the preceding tokens $x_{<i}$.

\vspace{-2.5mm}
\subsection{Results on Channels and Heads Pruning} \label{sec:HCPruning}
\vspace{-1.5mm}

The results of channels and heads pruning are shown in Table \ref{tab:channel_ppl_results}. Our method achieves the lowest perplexity scores. It verifies the superiority of optimization-based global and heterogeneous pruning. Especially, such outperformance is more significant at larger pruning rates over 40\%. The results on Mistral-7B-Instruct-v0.3 \cite{jiang2023mistral} are shown in Table \ref{tab:mistral} of Appendix \ref{app:mistral}. We further validate the method for pruning rates from 10\% to 50\% with more evaluation in Appendix \ref{app:more_eval}, and provide a comparison between our method and existing approaches that incorporate weight update, detailed in Appendix \ref{app:cmp_weight_update}.

\vspace{-2.6mm}
\subsection{Performance on Zero-shot Tasks} \label{sec:zero-shot}
\vspace{-1mm}

We follow SliceGPT \cite{ashkboos2024slicegpt} to assess our pruned LLM by EleutherAI LM Harness \cite{eval-harness} on five zero-shot tasks: PIQA \cite{bisk2020piqa}, WinoGrande \cite{sakaguchi2021winogrande}, HellaSwag \cite{zellers2019hellaswag}, ARC-e and ARC-c \cite{clark2018think} with the average scores across the five tasks. Our results on LLaMA-3-8B and LLaMA-2-7B in Tables \ref{tab:zero_shot_llama3} and \ref{tab:zero_shot_llama2} of Appendix \ref{app:lamma2}, demonstrate overall superior performance to the baselines, though C4-only pruning may negatively impact on particular cross-dataset zero-shot tasks such as Hellaswag \cite{zellers2019hellaswag}.

\begin{table*}[t]
\centering
\small
\begin{tabular}{@{}l||c|c||c|c||c|c@{}}
\hline
\hline
Method & PruneRate  & Perplexity & PruneRate  & Perplexity & PruneRate  & Perplexity  \\
\hline

LLM-Pruner & \multirow{4}{*}{30$\%$} & 38.94 & \multirow{4}{*}{40$\%$} & 68.48 & \multirow{4}{*}{50$\%$} & 190.56 \\
SliceGPT             & & 40.40 & & 73.76 & & 136.33 \\
Bonsai               & & 39.01 & & 69.18 & & 216.85 \\
Wanda-sp             & & 49.13 & & 78.45 & & 206.94 \\
\hline
\hline
Ours (Random Init)   & \multirow{2}{*}{30$\%$} & 37.24 & \multirow{2}{*}{40$\%$} & 60.16 & \multirow{2}{*}{50$\%$} & 160.75 \\
Ours (Random-Prog. Init)   & & \underline{31.43} & & \underline{49.86} & & \underline{86.55} \\
\hline
Ours (LLM-Pruner Init)   & \multirow{2}{*}{30$\%$} & 35.75 & \multirow{2}{*}{40$\%$} & 65.32 & \multirow{2}{*}{50$\%$} & 116.80 \\
Ours (Wanda-sp Init) & & \textbf{28.18} & & \textbf{39.81} & & \textbf{65.21} \\
\hline
\hline
\end{tabular}
\vspace{-3mm}
\caption{Channels and heads pruning results with \emph{different initializations} on LLaMA-2-7B. \textbf{Bold} and \underline{Underscored} denote the first and second best results, respectively.}
\label{tab:channel_diff_init}
\end{table*}

\begin{figure*}[t]
\vspace{-3mm}
\centering
\begin{minipage}[t]{0.67\textwidth}  
  \centering
  \includegraphics[width=\linewidth]{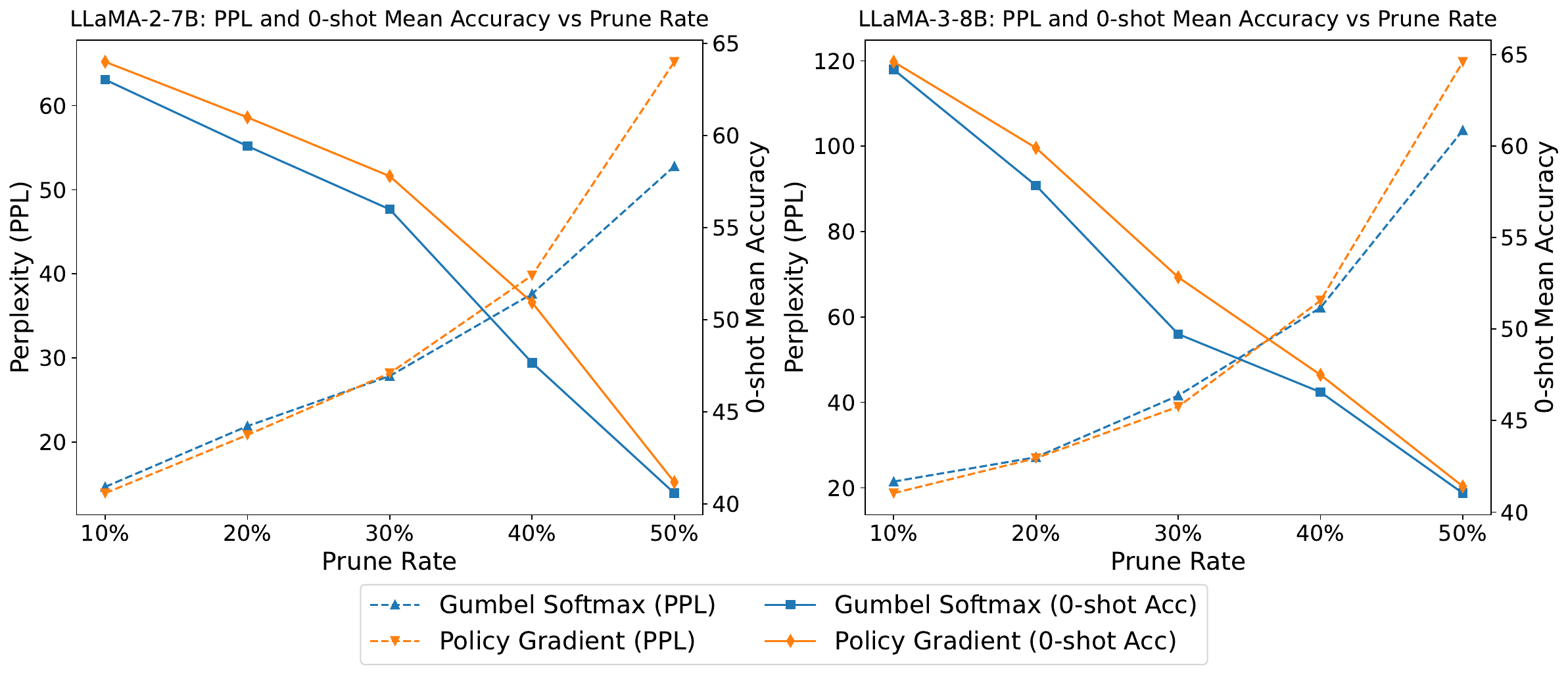} 
  \vspace{-7.5mm}
  \caption{Comparison of Policy Gradient and Gumbel Softmax.}
  \label{fig:gumbel_perform}
\end{minipage}%
\begin{minipage}[t]{.32\textwidth}  
  \centering
  \includegraphics[width=0.97\linewidth]{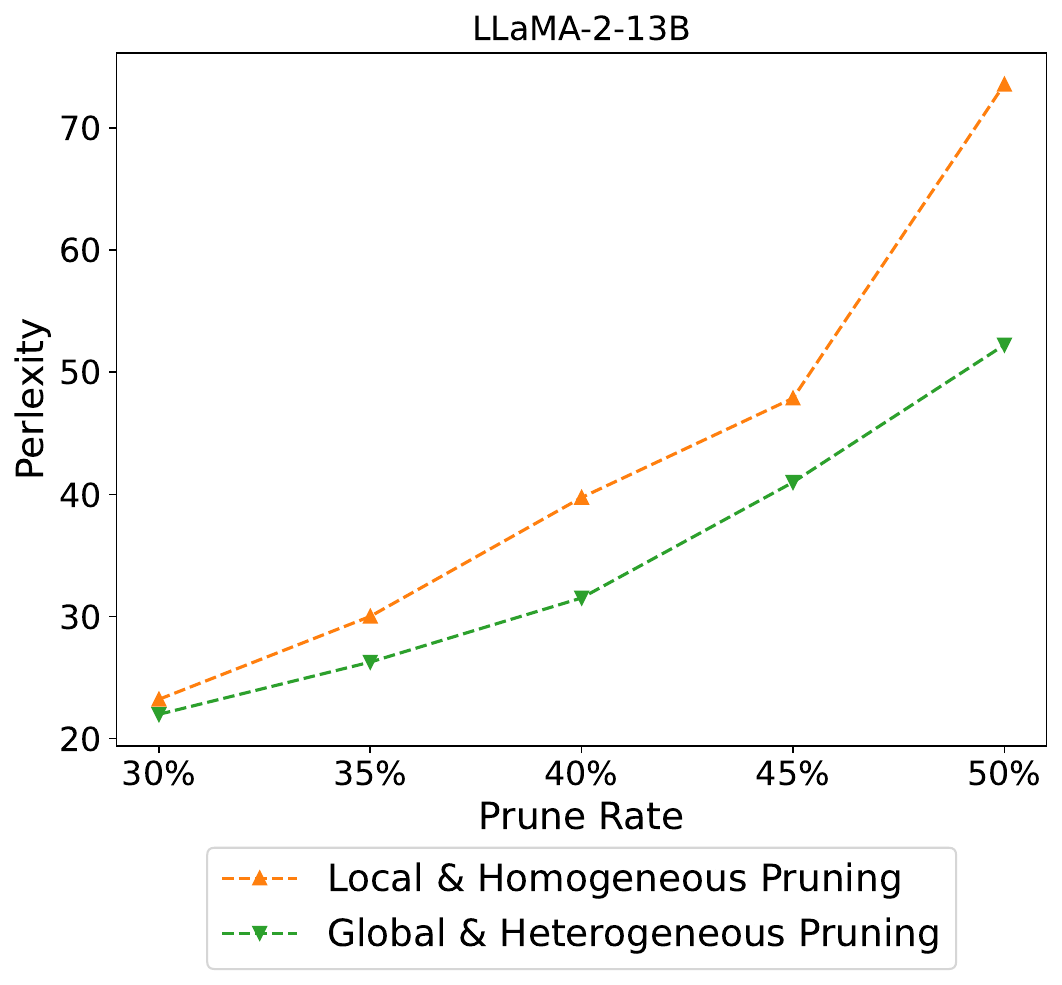}
  \vspace{-3mm}
  \caption{Global vs. local pruning.}
  \label{fig:13b_GorL_sparsity}
\end{minipage}
\vspace{-5mm}
\end{figure*}

\vspace{-2mm}
\section{Ablation Analysis} \label{chap:ablation}
\vspace{-2mm}

We investigate 1) the effect of various initialization of our method in Sect. \ref{sec:init}, Appendices \ref{app:diff_init} and \ref{app:proj_strategy}, 2) comparison with Gumbel Softmax, 3) performance of global and heterogeneous pruning versus that of local and homogenous pruning in Sect. \ref{sec:global}, 4) the remaining modules after pruning in Appendix \ref{app:post-prune}, and 5) the effect of the variance-reduced policy gradient in Appendix \ref{app:vrpg_ablation}.



\vspace{-2mm}
\subsection{Different Initializations} \label{sec:init}
\vspace{-1mm}

Our \texttt{Bernoulli} policy requires initialization to perform policy gradient optimization and to sample pruning masks. In this section, we investigate the \textbf{effect} and the \textbf{necessity} of using different metric-based methods as initializations. 
Moreover, the initialization of the \texttt{Bernoulli} policy should be probabilistic values between 0 and 1, but the metrics calculated by the metric-based methods \cite{sun2023simple, ma2023llm} may not hold this range. We thus discuss \textbf{different projection strategies} that transform those metrics to [0, 1] in Appendix \ref{app:proj_strategy}.

To address the practical case when a metric-based pruning is not \emph{apriori}, we propose progressive pruning with random initialization (\emph{Random-Progressive}), trained progressively with increasing pruning rates (each for only $1/3$ epoch). Details can be found in Appendix \ref{app:diff_init}. 

Different initializations are tested on LLaMA-2-7B. The baselines include simple random initialization with the target pruning rate (\emph{Random}) and progressive pruning with random initialization (\emph{Random-Progressive}). For channels and heads pruning, we investigate the initializations from Wanda-sp \cite{an2024fluctuation} and LLM-Pruner\footnote{We follow LLM-Pruner to fix the first four and the last two layers from pruning.} \cite{ma2023llm}, as shown in Table \ref{tab:channel_diff_init}. 

Our results in Tables \ref{tab:channel_diff_init} demonstrate that 1) different initializations lead to different results, 2) compared to the state-of-the-art methods, our method with most initializations except the random one exhibit new state-of-the-art results, and 3) The proposed \emph{Random-Progressive} initialization ranks the second place in most cases, surpassing previous state-of-the-art methods, \textbf{which suggests less necessity for employing a prior metric-based method to initiate our algorithm}.


\vspace{-2mm}
\subsection{Comparision with Gumbel Softmax} \label{sec:gumbel}
\vspace{-1mm}

As highlighted in Remark \ref{remark3}, our proposed PGE bypasses the costly back-propagation process through the LLM required by Gumbel Softmax, while maintaining comparable gradient estimation accuracy.

To substantiate this advantage, we conduct empirical ablation studies comparing different gradient estimators. The performance on LLaMA-2-7B and LLaMA-3-8B, measured by both perplexity and mean accuracy of 5 zero-shot tasks, of our PGE approach and back-propagation/Gumbel Softmax approach in Figure \ref{fig:gumbel_perform}. 
The performance of PGE is generally comparable to that of the Gumbel Softmax, except that PGE exhibits slightly higher perplexity at a 50\% pruning rate. This discrepancy may be attributed to the increased variance observed at this pruning level, which consequently amplifies the gradient estimation error.

We also illustrate the training time and memory usage of LLaMA-2-7B in Table \ref{tab:gumbel_time_mem}, which demonstrates that our method achieves comparable performance with significantly reduced resources. 

\begin{table}[hb]
\vspace{-2mm}
\centering
\footnotesize
\begin{tabular}{@{}c||c|c||c@{}}
\hline
\hline
\multirow{2}{*}{Method} & \multicolumn{2}{c||}{Memory (GiB)} & \multirow{2}{*}{Time (h)} \\ 
\cline{2-3}
                         & Min & Max &  \\
\hline
Gumbel Softmax           & 19.93 & 23.97 & 3.47 \\
\hline
Policy Gradient          & \textbf{17.23} & \textbf{17.39} & \textbf{1.56} \\
\hline
\hline
\end{tabular}
\vspace{-3mm}
\caption{Memory and Time Consumption Comparison between Gumbel Softmax and Policy Gradient.}
\label{tab:gumbel_time_mem}
\vspace{-5mm}
\end{table}

\vspace{-1mm}
\subsection{Merits of Global Pruning} \label{sec:global}
\vspace{-1mm}

Our method is able to perform global and heterogeneous pruning throughout the entire network, which is difficult for metric-based pruning methods \cite{sun2023simple, ma2023llm}, as the metrics across different layers often exhibit different magnitudes. As a compromise, those metric-based methods prune each layer locally and homogeneously. 

We validate the merits of global and heterogeneous pruning over local and homogeneous pruning, where we compare our method with a variant in which we prune each layer homogeneously. The channels and heads pruning results on LLaMA-2-13B are shown in Fig. \ref{fig:13b_GorL_sparsity}, demonstrating that the global and heterogeneous pruning significantly outperforms its local and homogeneous counterpart.

\vspace{-2mm}
\section{Conclusion} \label{sec:conclusion} 
\vspace{-2mm}

We propose an efficient optimization-based structural pruning method for LLMs, which 1) does not need back-propagation through the LLM \emph{per se}, 2) enables global and heterogeneous pruning throughout the LLM. Our method can take a metric-based pruning as initialization to achieve further improved performance. We implement our method by learning an underlying \texttt{Bernoulli} distribution of binary pruning mask. As we decouple the \texttt{Bernoulli} parameter and the sampled masks from the LLM loss, the \texttt{Bernoulli} distribution can thus be optimized by a policy gradient estimator without back-propagation through the LLM. Our method operates for 2.7 hours with approximately 35GB of memory on a single A100 GPU. Extensive experiments on various LLM models and sizes with detailed ablation analysis validate the promising performance of the proposed method.

\vspace{-2mm}
\section{Limitations} \label{sec:limitation}
\vspace{-2mm}

Firstly, as an optimization-based pruning, though our method exhibits \emph{improved performance} over the (heuristic) metric-based methods, and a \emph{similar memory complexity} (approximately 35GB, as only LLM forward is required), it simultaneously \emph{requires more training time} for optimization (\eg, 2.7 hours for LLaMA-2-13B) than the metric-based pruning methods.

Secondly, there exist advanced policy gradient algorithms with potentially lower variance from the reinforcement learning community. As 1) the primary focus of this paper is on the back-propagation-free formulation of the LLM pruning problem, and 2) our formulation ensures dense rewards at each step, we thus use a basic policy gradient algorithm similar to REINFORCE with simple variance reduction using a moving average baseline. We leave exploiting more powerful policy gradient algorithms as our future work.

Lastly, the performance of the proposed method on specific domains/tasks can rely heavily on the availability of domain-specific datasets. Though the cross-dataset evaluation is verified \wrt perplexity, pruning with only C4 dataset might have a negative influence on certain cross-dataset zero-shot tasks such as WinoGrande and Hellaswag.

\vspace{-2mm}
\section{Ethical Considerations} \label{sec:ethic}
\vspace{-1.5mm}

We have developed an efficient pruning method for Large Language Models (LLMs) that significantly accelerates inference speed. This approach optimizes computational efficiency and reduces energy consumption for online-deployed LLMs like ChatGPT, improving user experience while promoting sustainable AI. However, it also inherits the inherent ethical challenges of LLM technologies, requiring careful consideration in deployment.



\vspace{-2mm}
\section{Acknowledgements} \label{sec:acknowledge}
\vspace{-1.5mm}

This work was supported by the National Natural Science Foundation of China (62306214, 62325111, 62472097), the Natural Science Foundation of Hubei Province of China (2023AFB196), the Knowledge Innovation Program of Wuhan-Shuguang Project (2023010201020258), and Shanghai Municipal Science and Technology Commission (24511106102).

\bibliography{custom}

\begin{thebibliography}{78}
\providecommand{\natexlab}[1]{#1}

\bibitem[{Achiam et~al.(2023)Achiam, Adler, Agarwal, Ahmad, Akkaya, Aleman, Almeida, Altenschmidt, Altman, Anadkat et~al.}]{achiam2023gpt}
Josh Achiam, Steven Adler, Sandhini Agarwal, Lama Ahmad, Ilge Akkaya, Florencia~Leoni Aleman, Diogo Almeida, Janko Altenschmidt, Sam Altman, Shyamal Anadkat, et~al. 2023.
\newblock Gpt-4 technical report.
\newblock \emph{arXiv preprint arXiv:2303.08774}.

\bibitem[{An et~al.(2024)An, Zhao, Yu, Tang, and Wang}]{an2024fluctuation}
Yongqi An, Xu~Zhao, Tao Yu, Ming Tang, and Jinqiao Wang. 2024.
\newblock Fluctuation-based adaptive structured pruning for large language models.
\newblock In \emph{Proceedings of the AAAI Conference on Artificial Intelligence}, volume~38, pages 10865--10873.

\bibitem[{Ashkboos et~al.(2024)Ashkboos, Croci, Nascimento, Hoefler, and Hensman}]{ashkboos2024slicegpt}
Saleh Ashkboos, Maximilian~L Croci, Marcelo Gennari~do Nascimento, Torsten Hoefler, and James Hensman. 2024.
\newblock Slicegpt: Compress large language models by deleting rows and columns.
\newblock \emph{arXiv preprint arXiv:2401.15024}.

\bibitem[{Bai et~al.(2024)Bai, Li, Ling, Kim, and Zhao}]{bai2024sparsellm}
Guangji Bai, Yijiang Li, Chen Ling, Kibaek Kim, and Liang Zhao. 2024.
\newblock Sparsellm: Towards global pruning of pre-trained language models.
\newblock In \emph{The Thirty-eighth Annual Conference on Neural Information Processing Systems}.

\bibitem[{Bisk et~al.(2020)Bisk, Zellers, Gao, Choi et~al.}]{bisk2020piqa}
Yonatan Bisk, Rowan Zellers, Jianfeng Gao, Yejin Choi, et~al. 2020.
\newblock Piqa: Reasoning about physical commonsense in natural language.
\newblock In \emph{AAAI}, volume~34, pages 7432--7439.

\bibitem[{Blalock et~al.(2020)Blalock, Gonzalez~Ortiz, Frankle, and Guttag}]{blalock2020state}
Davis Blalock, Jose~Javier Gonzalez~Ortiz, Jonathan Frankle, and John Guttag. 2020.
\newblock What is the state of neural network pruning?
\newblock \emph{Proceedings of machine learning and systems}, 2:129--146.

\bibitem[{Bo{\v{z}}a(2024)}]{bovza2024fast}
Vladim{\'\i}r Bo{\v{z}}a. 2024.
\newblock Fast and optimal weight update for pruned large language models.
\newblock \emph{arXiv preprint arXiv:2401.02938}.

\bibitem[{Brown et~al.(2020)Brown, Mann, Ryder, Subbiah, Kaplan, Dhariwal, Neelakantan, Shyam, Sastry, Askell et~al.}]{brown2020language}
Tom Brown, Benjamin Mann, Nick Ryder, Melanie Subbiah, Jared~D Kaplan, Prafulla Dhariwal, Arvind Neelakantan, Pranav Shyam, Girish Sastry, Amanda Askell, et~al. 2020.
\newblock Language models are few-shot learners.
\newblock In \emph{NeurIPS}, pages 1877--1901.

\bibitem[{Chen et~al.(2023)Chen, Ding, Yadav, Zharkov, and Liang}]{chen2023lorashear}
Tianyi Chen, Tianyu Ding, Badal Yadav, Ilya Zharkov, and Luming Liang. 2023.
\newblock Lorashear: Efficient large language model structured pruning and knowledge recovery.
\newblock \emph{arXiv preprint arXiv:2310.18356}.

\bibitem[{Chen et~al.(2024)Chen, Hu, and Zhang}]{chen2024compressing}
Xiaodong Chen, Yuxuan Hu, and Jing Zhang. 2024.
\newblock Compressing large language models by streamlining the unimportant layer.
\newblock \emph{arXiv preprint arXiv:2403.19135}.

\bibitem[{Chiang et~al.(2023)Chiang, Li, Lin, Sheng, Wu, Zhang, Zheng, Zhuang, Zhuang, Gonzalez, Stoica, and Xing}]{chiang2023vicuna}
Wei-Lin Chiang, Zhuohan Li, Zi~Lin, Ying Sheng, Zhanghao Wu, Hao Zhang, Lianmin Zheng, Siyuan Zhuang, Yonghao Zhuang, Joseph~E. Gonzalez, Ion Stoica, and Eric~P. Xing. 2023.
\newblock \href {https://lmsys.org/blog/2023-03-30-vicuna/} {Vicuna: An open-source chatbot impressing gpt-4 with 90\%* chatgpt quality}.

\bibitem[{Clark et~al.(2019)Clark, Lee, Chang, Kwiatkowski, Collins, and Toutanova}]{clark2019boolq}
Christopher Clark, Kenton Lee, Ming-Wei Chang, Tom Kwiatkowski, Michael Collins, and Kristina Toutanova. 2019.
\newblock Boolq: Exploring the surprising difficulty of natural yes/no questions.
\newblock \emph{arXiv preprint arXiv:1905.10044}.

\bibitem[{Clark et~al.(2018)Clark, Cowhey, Etzioni, Khot, Sabharwal, Schoenick, and Tafjord}]{clark2018think}
Peter Clark, Isaac Cowhey, Oren Etzioni, Tushar Khot, Ashish Sabharwal, Carissa Schoenick, and Oyvind Tafjord. 2018.
\newblock Think you have solved question answering? try arc, the ai2 reasoning challenge.
\newblock \emph{arXiv preprint arXiv:1803.05457}.

\bibitem[{Das et~al.(2023)Das, Ma, and Shen}]{das2023beyond}
Rocktim~Jyoti Das, Liqun Ma, and Zhiqiang Shen. 2023.
\newblock Beyond size: How gradients shape pruning decisions in large language models.
\newblock \emph{arXiv preprint arXiv:2311.04902}.

\bibitem[{Dery et~al.(2024)Dery, Kolawole, Kagey, Smith, Neubig, and Talwalkar}]{dery2024everybody}
Lucio Dery, Steven Kolawole, Jean-Francois Kagey, Virginia Smith, Graham Neubig, and Ameet Talwalkar. 2024.
\newblock Everybody prune now: Structured pruning of llms with only forward passes.
\newblock \emph{arXiv preprint arXiv:2402.05406}.

\bibitem[{Dubey et~al.(2024)Dubey, Jauhri, Pandey, Kadian, Al-Dahle, Letman, Mathur, Schelten, Yang, Fan et~al.}]{dubey2024llama}
Abhimanyu Dubey, Abhinav Jauhri, Abhinav Pandey, Abhishek Kadian, Ahmad Al-Dahle, Aiesha Letman, Akhil Mathur, Alan Schelten, Amy Yang, Angela Fan, et~al. 2024.
\newblock The llama 3 herd of models.
\newblock \emph{arXiv preprint arXiv:2407.21783}.

\bibitem[{Dupont et~al.(2022)Dupont, Alaoui, Sahbi, and Lebois}]{dupont2022extracting}
Robin Dupont, Mohammed~Amine Alaoui, Hichem Sahbi, and Alice Lebois. 2022.
\newblock Extracting effective subnetworks with gumbel-softmax.
\newblock In \emph{ICIP}, pages 931--935.

\bibitem[{Evci et~al.(2020)Evci, Gale, Menick, Castro, and Elsen}]{evci2020rigging}
Utku Evci, Trevor Gale, Jacob Menick, Pablo~Samuel Castro, and Erich Elsen. 2020.
\newblock Rigging the lottery: Making all tickets winners.
\newblock In \emph{International conference on machine learning}, pages 2943--2952. PMLR.

\bibitem[{Fang et~al.(2024)Fang, Yin, Muralidharan, Heinrich, Pool, Kautz, Molchanov, and Wang}]{fang2024maskllm}
Gongfan Fang, Hongxu Yin, Saurav Muralidharan, Greg Heinrich, Jeff Pool, Jan Kautz, Pavlo Molchanov, and Xinchao Wang. 2024.
\newblock Maskllm: Learnable semi-structured sparsity for large language models.
\newblock \emph{arXiv preprint arXiv:2409.17481}.

\bibitem[{Frankle and Carbin(2018)}]{frankle2018lottery}
Jonathan Frankle and Michael Carbin. 2018.
\newblock The lottery ticket hypothesis: Finding sparse, trainable neural networks.
\newblock \emph{arXiv preprint arXiv:1803.03635}.

\bibitem[{Frantar and Alistarh(2023)}]{frantar2023sparsegpt}
Elias Frantar and Dan Alistarh. 2023.
\newblock Sparsegpt: Massive language models can be accurately pruned in one-shot.
\newblock In \emph{International Conference on Machine Learning}, pages 10323--10337. PMLR.

\bibitem[{Gao et~al.(2023)Gao, Tow, Abbasi, Biderman, Black, DiPofi, Foster, Golding, Hsu, Le~Noac'h, Li, McDonell, Muennighoff, Ociepa, Phang, Reynolds, Schoelkopf, Skowron, Sutawika, Tang, Thite, Wang, Wang, and Zou}]{eval-harness}
Leo Gao, Jonathan Tow, Baber Abbasi, Stella Biderman, Sid Black, Anthony DiPofi, Charles Foster, Laurence Golding, Jeffrey Hsu, Alain Le~Noac'h, Haonan Li, Kyle McDonell, Niklas Muennighoff, Chris Ociepa, Jason Phang, Laria Reynolds, Hailey Schoelkopf, Aviya Skowron, Lintang Sutawika, Eric Tang, Anish Thite, Ben Wang, Kevin Wang, and Andy Zou. 2023.
\newblock \href {https://doi.org/10.5281/zenodo.10256836} {A framework for few-shot language model evaluation}.

\bibitem[{Guo et~al.(2023)Guo, Xu, Zhang, and Yang}]{guo2023compresso}
Song Guo, Jiahang Xu, Li~Lyna Zhang, and Mao Yang. 2023.
\newblock Compresso: Structured pruning with collaborative prompting learns compact large language models.
\newblock \emph{arXiv preprint arXiv:2310.05015}.

\bibitem[{Han et~al.(2015)Han, Mao, and Dally}]{han2015deep}
Song Han, Huizi Mao, and William~J Dally. 2015.
\newblock Deep compression: Compressing deep neural networks with pruning, trained quantization and huffman coding.
\newblock \emph{arXiv preprint arXiv:1510.00149}.

\bibitem[{He et~al.(2018)He, Lin, Liu, Wang, Li, and Han}]{he2018amc}
Yihui He, Ji~Lin, Zhijian Liu, Hanrui Wang, Li-Jia Li, and Song Han. 2018.
\newblock Amc: Automl for model compression and acceleration on mobile devices.
\newblock In \emph{ECCV}, pages 784--800.

\bibitem[{Hendrycks et~al.(2020)Hendrycks, Burns, Basart, Zou, Mazeika, Song, and Steinhardt}]{hendrycks2020measuring}
Dan Hendrycks, Collin Burns, Steven Basart, Andy Zou, Mantas Mazeika, Dawn Song, and Jacob Steinhardt. 2020.
\newblock Measuring massive multitask language understanding.
\newblock \emph{arXiv preprint arXiv:2009.03300}.

\bibitem[{Hu et~al.(2022)Hu, Shen, Wallis, Allen-Zhu, Li, Wang, Wang, and Chen}]{hu2021lora}
Edward~J Hu, Yelong Shen, Phillip Wallis, Zeyuan Allen-Zhu, Yuanzhi Li, Shean Wang, Lu~Wang, and Weizhu Chen. 2022.
\newblock Lora: Low-rank adaptation of large language models.
\newblock In \emph{ICLR}.

\bibitem[{Huang and Wang(2018)}]{huang2018data}
Zehao Huang and Naiyan Wang. 2018.
\newblock Data-driven sparse structure selection for deep neural networks.
\newblock In \emph{Proceedings of the European conference on computer vision (ECCV)}, pages 304--320.

\bibitem[{Huijben et~al.(2022)Huijben, Kool, Paulus, and Van~Sloun}]{huijben2022review}
Iris~AM Huijben, Wouter Kool, Max~B Paulus, and Ruud~JG Van~Sloun. 2022.
\newblock A review of the gumbel-max trick and its extensions for discrete stochasticity in machine learning.
\newblock \emph{IEEE TPAMI}, 45(2):1353--1371.

\bibitem[{Jiang et~al.(2023)Jiang, Sablayrolles, Mensch, Bamford, Chaplot, Casas, Bressand, Lengyel, Lample, Saulnier et~al.}]{jiang2023mistral}
Albert~Q Jiang, Alexandre Sablayrolles, Arthur Mensch, Chris Bamford, Devendra~Singh Chaplot, Diego de~las Casas, Florian Bressand, Gianna Lengyel, Guillaume Lample, Lucile Saulnier, et~al. 2023.
\newblock Mistral 7b.
\newblock \emph{arXiv preprint arXiv:2310.06825}.

\bibitem[{Kim et~al.(2024)Kim, Kim, Kim, Castells, Choi, Shin, and Song}]{kim2024shortened}
Bo-Kyeong Kim, Geonmin Kim, Tae-Ho Kim, Thibault Castells, Shinkook Choi, Junho Shin, and Hyoung-Kyu Song. 2024.
\newblock Shortened llama: A simple depth pruning for large language models.
\newblock \emph{arXiv preprint arXiv:2402.02834}.

\bibitem[{Ko et~al.(2023)Ko, Park, Kim, Ahn, Chang, Ahn, and Yun}]{ko2023nash}
Jongwoo Ko, Seungjoon Park, Yujin Kim, Sumyeong Ahn, Du-Seong Chang, Euijai Ahn, and Se-Young Yun. 2023.
\newblock Nash: A simple unified framework of structured pruning for accelerating encoder-decoder language models.
\newblock \emph{arXiv preprint arXiv:2310.10054}.

\bibitem[{Kurtic et~al.(2022)Kurtic, Campos, Nguyen, Frantar, Kurtz, Fineran, Goin, and Alistarh}]{kurtic2022optimal}
Eldar Kurtic, Daniel Campos, Tuan Nguyen, Elias Frantar, Mark Kurtz, Benjamin Fineran, Michael Goin, and Dan Alistarh. 2022.
\newblock The optimal bert surgeon: Scalable and accurate second-order pruning for large language models.
\newblock In \emph{Proceedings of the 2022 Conference on Empirical Methods in Natural Language Processing}, pages 4163--4181.

\bibitem[{Lai et~al.(2017)Lai, Xie, Liu, Yang, and Hovy}]{lai2017race}
Guokun Lai, Qizhe Xie, Hanxiao Liu, Yiming Yang, and Eduard Hovy. 2017.
\newblock Race: Large-scale reading comprehension dataset from examinations.
\newblock \emph{arXiv preprint arXiv:1704.04683}.

\bibitem[{Li et~al.(2024{\natexlab{a}})Li, Dong, Tang, Liu, Wang, Luo, Xue, Liu, Chu, and Guo}]{li2024discovering}
Lujun Li, Peijie Dong, Zhenheng Tang, Xiang Liu, Qiang Wang, Wenhan Luo, Wei Xue, Qifeng Liu, Xiaowen Chu, and Yike Guo. 2024{\natexlab{a}}.
\newblock Discovering sparsity allocation for layer-wise pruning of large language models.
\newblock In \emph{The Thirty-eighth Annual Conference on Neural Information Processing Systems}.

\bibitem[{Li et~al.(2024{\natexlab{b}})Li, Han, and Bai}]{li2024nuteprune}
Shengrui Li, Xueting Han, and Jing Bai. 2024{\natexlab{b}}.
\newblock Nuteprune: Efficient progressive pruning with numerous teachers for large language models.
\newblock \emph{arXiv preprint arXiv:2402.09773}.

\bibitem[{Li et~al.(2023)Li, Niu, Zhang, Liu, Zhu, and Kang}]{li2023sparse}
Yun Li, Lin Niu, Xipeng Zhang, Kai Liu, Jianchen Zhu, and Zhanhui Kang. 2023.
\newblock E-sparse: Boosting the large language model inference through entropy-based {N}: {M} sparsity.
\newblock \emph{arXiv preprint arXiv:2310.15929}.

\bibitem[{Liu et~al.(2018{\natexlab{a}})Liu, Simonyan, and Yang}]{liu2018darts}
Hanxiao Liu, Karen Simonyan, and Yiming Yang. 2018{\natexlab{a}}.
\newblock Darts: Differentiable architecture search.
\newblock \emph{arXiv preprint arXiv:1806.09055}.

\bibitem[{Liu et~al.(2020)Liu, Zhang, Basar, and Yin}]{liu2020improved}
Yanli Liu, Kaiqing Zhang, Tamer Basar, and Wotao Yin. 2020.
\newblock An improved analysis of (variance-reduced) policy gradient and natural policy gradient methods.
\newblock \emph{NeurIPS}, 33:7624--7636.

\bibitem[{Liu et~al.(2019)Liu, Mu, Zhang, Guo, Yang, Cheng, and Sun}]{liu2019metapruning}
Zechun Liu, Haoyuan Mu, Xiangyu Zhang, Zichao Guo, Xin Yang, Kwang-Ting Cheng, and Jian Sun. 2019.
\newblock Metapruning: Meta learning for automatic neural network channel pruning.
\newblock In \emph{ICCV}.

\bibitem[{Liu et~al.(2018{\natexlab{b}})Liu, Sun, Zhou, Huang, and Darrell}]{liu2018rethinking}
Zhuang Liu, Mingjie Sun, Tinghui Zhou, Gao Huang, and Trevor Darrell. 2018{\natexlab{b}}.
\newblock Rethinking the value of network pruning.
\newblock \emph{arXiv preprint arXiv:1810.05270}.

\bibitem[{Lu et~al.(2024)Lu, Zhou, Liu, Wang, Mahoney, and Yang}]{lu2024alphapruning}
Haiquan Lu, Yefan Zhou, Shiwei Liu, Zhangyang Wang, Michael~W Mahoney, and Yaoqing Yang. 2024.
\newblock Alphapruning: Using heavy-tailed self regularization theory for improved layer-wise pruning of large language models.
\newblock \emph{arXiv preprint arXiv:2410.10912}.

\bibitem[{Ma et~al.(2023)Ma, Fang, and Wang}]{ma2023llm}
Xinyin Ma, Gongfan Fang, and Xinchao Wang. 2023.
\newblock Llm-pruner: On the structural pruning of large language models.
\newblock \emph{Advances in neural information processing systems}, 36:21702--21720.

\bibitem[{Maddison et~al.(2016)Maddison, Mnih, and Teh}]{maddison2016concrete}
Chris~J Maddison, Andriy Mnih, and Yee~Whye Teh. 2016.
\newblock The concrete distribution: A continuous relaxation of discrete random variables.
\newblock \emph{arXiv preprint arXiv:1611.00712}.

\bibitem[{Mangrulkar et~al.(2022)Mangrulkar, Gugger, Debut, Belkada, Paul, and Bossan}]{mangrulkar2022peft}
Sourab Mangrulkar, Sylvain Gugger, Lysandre Debut, Younes Belkada, Sayak Paul, and B~Bossan. 2022.
\newblock \href {https://github. com/huggingface/peft} {Peft: State-of-the-art parameter-efficient fine-tuning methods}.

\bibitem[{Meng et~al.(2024)Meng, Behdin, Wang, and Mazumder}]{meng2024alps}
Xiang Meng, Kayhan Behdin, Haoyue Wang, and Rahul Mazumder. 2024.
\newblock Alps: Improved optimization for highly sparse one-shot pruning for large language models.
\newblock \emph{arXiv preprint arXiv:2406.07831}.

\bibitem[{Merity et~al.(2016)Merity, Xiong, Bradbury, and Socher}]{merity2016pointer}
Stephen Merity, Caiming Xiong, James Bradbury, and Richard Socher. 2016.
\newblock Pointer sentinel mixture models.
\newblock \emph{arXiv preprint arXiv:1609.07843}.

\bibitem[{Molchanov et~al.(2019)Molchanov, Mallya, Tyree, Frosio, and Kautz}]{molchanov2019importance}
Pavlo Molchanov, Arun Mallya, Stephen Tyree, Iuri Frosio, and Jan Kautz. 2019.
\newblock Importance estimation for neural network pruning.
\newblock In \emph{Proceedings of the IEEE/CVF conference on computer vision and pattern recognition}, pages 11264--11272.

\bibitem[{Muralidharan et~al.(2024)Muralidharan, Sreenivas, Joshi, Chochowski, Patwary, Shoeybi, Catanzaro, Kautz, and Molchanov}]{muralidharan2024compact}
Saurav Muralidharan, Sharath~Turuvekere Sreenivas, Raviraj~Bhuminand Joshi, Marcin Chochowski, Mostofa Patwary, Mohammad Shoeybi, Bryan Catanzaro, Jan Kautz, and Pavlo Molchanov. 2024.
\newblock Compact language models via pruning and knowledge distillation.
\newblock In \emph{The Thirty-eighth Annual Conference on Neural Information Processing Systems}.

\bibitem[{Paperno et~al.(2016)Paperno, Kruszewski, Lazaridou, Pham, Bernardi, Pezzelle, Baroni, Boleda, and Fern{\'a}ndez}]{paperno2016lambada}
Denis Paperno, Germ{\'a}n Kruszewski, Angeliki Lazaridou, Quan~Ngoc Pham, Raffaella Bernardi, Sandro Pezzelle, Marco Baroni, Gemma Boleda, and Raquel Fern{\'a}ndez. 2016.
\newblock The lambada dataset: Word prediction requiring a broad discourse context.
\newblock \emph{arXiv preprint arXiv:1606.06031}.

\bibitem[{Raffel et~al.(2020)Raffel, Shazeer, Roberts, Lee, Narang, Matena, Zhou, Li, and Liu}]{raffel2020exploring}
Colin Raffel, Noam Shazeer, Adam Roberts, Katherine Lee, Sharan Narang, Michael Matena, Yanqi Zhou, Wei Li, and Peter~J Liu. 2020.
\newblock Exploring the limits of transfer learning with a unified text-to-text transformer.
\newblock \emph{JMLR}, 21(140):1--67.

\bibitem[{Sakaguchi et~al.(2021)Sakaguchi, Bras, Bhagavatula, and Choi}]{sakaguchi2021winogrande}
Keisuke Sakaguchi, Ronan~Le Bras, Chandra Bhagavatula, and Yejin Choi. 2021.
\newblock Winogrande: An adversarial winograd schema challenge at scale.
\newblock \emph{Communications of the ACM}, 64(9):99--106.

\bibitem[{Sehnke et~al.(2010)Sehnke, Osendorfer, R{\"u}ckstie{\ss}, Graves, Peters, and Schmidhuber}]{sehnke2010parameter}
Frank Sehnke, Christian Osendorfer, Thomas R{\"u}ckstie{\ss}, Alex Graves, Jan Peters, and J{\"u}rgen Schmidhuber. 2010.
\newblock Parameter-exploring policy gradients.
\newblock \emph{Neural Networks}, 23(4):551--559.

\bibitem[{Shao et~al.(2024)Shao, Liu, and Qian}]{shao2024one}
Hang Shao, Bei Liu, and Yanmin Qian. 2024.
\newblock One-shot sensitivity-aware mixed sparsity pruning for large language models.
\newblock In \emph{ICASSP 2024-2024 IEEE International Conference on Acoustics, Speech and Signal Processing (ICASSP)}, pages 11296--11300. IEEE.

\bibitem[{Shen et~al.(2024)Shen, Zhao, Gong, Kong, Zhan, Wu, Lin, Wu, Lin, and Wang}]{shen2024search}
Xuan Shen, Pu~Zhao, Yifan Gong, Zhenglun Kong, Zheng Zhan, Yushu Wu, Ming Lin, Chao Wu, Xue Lin, and Yanzhi Wang. 2024.
\newblock Search for efficient large language models.
\newblock \emph{arXiv preprint arXiv:2409.17372}.

\bibitem[{Song et~al.(2024)Song, Oh, Kim, Kim, Kim, and Kim}]{song2024sleb}
Jiwon Song, Kyungseok Oh, Taesu Kim, Hyungjun Kim, Yulhwa Kim, and Jae-Joon Kim. 2024.
\newblock Sleb: Streamlining llms through redundancy verification and elimination of transformer blocks.
\newblock \emph{arXiv preprint arXiv:2402.09025}.

\bibitem[{Srinivas et~al.(2017)Srinivas, Subramanya, and Venkatesh~Babu}]{srinivas2017training}
Suraj Srinivas, Akshayvarun Subramanya, and R~Venkatesh~Babu. 2017.
\newblock Training sparse neural networks.
\newblock In \emph{CVPR Workshops}, pages 138--145.

\bibitem[{Sun et~al.(2023)Sun, Liu, Bair, and Kolter}]{sun2023simple}
Mingjie Sun, Zhuang Liu, Anna Bair, and J~Zico Kolter. 2023.
\newblock A simple and effective pruning approach for large language models.
\newblock \emph{arXiv preprint arXiv:2306.11695}.

\bibitem[{Tan et~al.(2024)Tan, Zhang, and Wei}]{tan2024wrp}
Zhendong Tan, Xingjun Zhang, and Zheng Wei. 2024.
\newblock Wrp: Weight recover prune for structured sparsity.
\newblock In \emph{Proceedings of the 62nd Annual Meeting of the Association for Computational Linguistics (Volume 1: Long Papers)}, pages 6433--6443.

\bibitem[{Taori et~al.(2023)Taori, Gulrajani, Zhang, Dubois, Li, Guestrin, Liang, and Hashimoto}]{taori2023stanford}
Rohan Taori, Ishaan Gulrajani, Tianyi Zhang, Yann Dubois, Xuechen Li, Carlos Guestrin, Percy Liang, and Tatsunori~B Hashimoto. 2023.
\newblock Stanford alpaca: An instruction-following llama model.

\bibitem[{Touvron et~al.(2023{\natexlab{a}})Touvron, Lavril, Izacard, Martinet, Lachaux, Lacroix, Rozi{\`e}re, Goyal, Hambro, Azhar et~al.}]{touvron2023llama1}
Hugo Touvron, Thibaut Lavril, Gautier Izacard, Xavier Martinet, Marie-Anne Lachaux, Timoth{\'e}e Lacroix, Baptiste Rozi{\`e}re, Naman Goyal, Eric Hambro, Faisal Azhar, et~al. 2023{\natexlab{a}}.
\newblock Llama: Open and efficient foundation language models.
\newblock \emph{arXiv preprint arXiv:2302.13971}.

\bibitem[{Touvron et~al.(2023{\natexlab{b}})Touvron, Martin, Stone, Albert, Almahairi, Babaei, Bashlykov, Batra, Bhargava, Bhosale et~al.}]{touvron2023llama2}
Hugo Touvron, Louis Martin, Kevin Stone, Peter Albert, Amjad Almahairi, Yasmine Babaei, Nikolay Bashlykov, Soumya Batra, Prajjwal Bhargava, Shruti Bhosale, et~al. 2023{\natexlab{b}}.
\newblock Llama 2: Open foundation and fine-tuned chat models.
\newblock \emph{arXiv preprint arXiv:2307.09288}.

\bibitem[{van~der Ouderaa et~al.(2023)van~der Ouderaa, Nagel, van Baalen, Asano, and Blankevoort}]{van2023llm}
Tycho~FA van~der Ouderaa, Markus Nagel, Mart van Baalen, Yuki~M Asano, and Tijmen Blankevoort. 2023.
\newblock The llm surgeon.
\newblock \emph{arXiv preprint arXiv:2312.17244}.

\bibitem[{Wang et~al.(2021)Wang, Qin, Zhang, and Fu}]{wang2021neural}
Huan Wang, Can Qin, Yulun Zhang, and Yun Fu. 2021.
\newblock Neural pruning via growing regularization.
\newblock In \emph{ICLR}.

\bibitem[{Wang and Carreira-Perpin{\'a}n(2013)}]{wang2013projection}
Weiran Wang and Miguel~A Carreira-Perpin{\'a}n. 2013.
\newblock Projection onto the probability simplex: An efficient algorithm with a simple proof, and an application.
\newblock \emph{arXiv preprint arXiv:1309.1541}.

\bibitem[{Wei et~al.(2024)Wei, Huang, Huang, Xie, Qi, Xia, Mittal, Wang, and Henderson}]{wei2024assessing}
Boyi Wei, Kaixuan Huang, Yangsibo Huang, Tinghao Xie, Xiangyu Qi, Mengzhou Xia, Prateek Mittal, Mengdi Wang, and Peter Henderson. 2024.
\newblock Assessing the brittleness of safety alignment via pruning and low-rank modifications.
\newblock \emph{arXiv preprint arXiv:2402.05162}.

\bibitem[{Williams(1992)}]{williams1992simple}
Ronald~J Williams. 1992.
\newblock Simple statistical gradient-following algorithms for connectionist reinforcement learning.
\newblock \emph{Machine learning}, 8:229--256.

\bibitem[{Xia et~al.(2023)Xia, Gao, Zeng, and Chen}]{xia2023sheared}
Mengzhou Xia, Tianyu Gao, Zhiyuan Zeng, and Danqi Chen. 2023.
\newblock Sheared llama: Accelerating language model pre-training via structured pruning.
\newblock \emph{arXiv preprint arXiv:2310.06694}.

\bibitem[{Xu et~al.(2024)Xu, Shao, Chen, Tang, Zhang, Gao, An, Qiao, and Luo}]{xu2024besa}
Peng Xu, Wenqi Shao, Mengzhao Chen, Shitao Tang, Kaipeng Zhang, Peng Gao, Fengwei An, Yu~Qiao, and Ping Luo. 2024.
\newblock Besa: Pruning large language models with blockwise parameter-efficient sparsity allocation.
\newblock \emph{arXiv preprint arXiv:2402.16880}.

\bibitem[{Yang et~al.(2024)Yang, Cao, and Zhao}]{yang2024laco}
Yifei Yang, Zouying Cao, and Hai Zhao. 2024.
\newblock Laco: Large language model pruning via layer collapse.
\newblock \emph{arXiv preprint arXiv:2402.11187}.

\bibitem[{Yin et~al.(2023)Yin, Wu, Zhang, Hsieh, Wang, Jia, Pechenizkiy, Liang, Wang, and Liu}]{yin2023outlier}
Lu~Yin, You Wu, Zhenyu Zhang, Cheng-Yu Hsieh, Yaqing Wang, Yiling Jia, Mykola Pechenizkiy, Yi~Liang, Zhangyang Wang, and Shiwei Liu. 2023.
\newblock Outlier weighed layerwise sparsity (owl): A missing secret sauce for pruning llms to high sparsity.
\newblock \emph{arXiv preprint arXiv:2310.05175}.

\bibitem[{Zellers et~al.(2019)Zellers, Holtzman, Bisk, Farhadi, and Choi}]{zellers2019hellaswag}
Rowan Zellers, Ari Holtzman, Yonatan Bisk, Ali Farhadi, and Yejin Choi. 2019.
\newblock Hellaswag: Can a machine really finish your sentence?
\newblock \emph{arXiv preprint arXiv:1905.07830}.

\bibitem[{Zeng et~al.(2024)Zeng, Xu, Chen, and Yu}]{zeng2024multilingual}
Hongchuan Zeng, Hongshen Xu, Lu~Chen, and Kai Yu. 2024.
\newblock Multilingual brain surgeon: Large language models can be compressed leaving no language behind.
\newblock \emph{arXiv preprint arXiv:2404.04748}.

\bibitem[{Zhang et~al.(2024{\natexlab{a}})Zhang, Chen, Shen, Yang, Ou, Yu, and Zhuang}]{zhang2024loraprune}
Mingyang Zhang, Hao Chen, Chunhua Shen, Zhen Yang, Linlin Ou, Xinyi Yu, and Bohan Zhuang. 2024{\natexlab{a}}.
\newblock Loraprune: Structured pruning meets low-rank parameter-efficient fine-tuning.
\newblock In \emph{Findings of the Association for Computational Linguistics ACL 2024}, pages 3013--3026.

\bibitem[{Zhang et~al.(2024{\natexlab{b}})Zhang, Bai, Lin, Zhao, Hou, and Cannistraci}]{zhang2023efficient}
Yingtao Zhang, Haoli Bai, Haokun Lin, Jialin Zhao, Lu~Hou, and Carlo~Vittorio Cannistraci. 2024{\natexlab{b}}.
\newblock Plug-and-play: An efficient post-training pruning method for large language models.
\newblock In \emph{ICLR}.

\bibitem[{Zhang et~al.(2023)Zhang, Zhao, Lin, Sun, Yao, Han, Tanner, Liu, and Ji}]{zhang2023dynamic}
Yuxin Zhang, Lirui Zhao, Mingbao Lin, Yunyun Sun, Yiwu Yao, Xingjia Han, Jared Tanner, Shiwei Liu, and Rongrong Ji. 2023.
\newblock Dynamic sparse no training: Training-free fine-tuning for sparse llms.
\newblock \emph{arXiv preprint arXiv:2310.08915}.

\bibitem[{Zhao et~al.(2024)Zhao, Hajishirzi, and Cao}]{zhao2024apt}
Bowen Zhao, Hannaneh Hajishirzi, and Qingqing Cao. 2024.
\newblock Apt: Adaptive pruning and tuning pretrained language models for efficient training and inference.
\newblock \emph{arXiv preprint arXiv:2401.12200}.

\bibitem[{Zhao et~al.(2011)Zhao, Hachiya, Niu, and Sugiyama}]{zhao2011analysis}
Tingting Zhao, Hirotaka Hachiya, Gang Niu, and Masashi Sugiyama. 2011.
\newblock Analysis and improvement of policy gradient estimation.
\newblock In \emph{NIPS}.

\end{thebibliography}




\appendix
\renewcommand{\thefigure}{A\arabic{figure}}
\renewcommand{\thetable}{A\arabic{table}}
\renewcommand{\theequation}{A\arabic{equation}}
\setcounter{figure}{0}
\setcounter{table}{0}
\setcounter{equation}{0}

\begin{table*}[ht]
\centering
\small
\begin{tabular}{|p{0.08\textwidth}|p{0.86\textwidth}|}
\hline
\hline

\multirow{4}{*}{\textbf{Analysis}} & \makecell[l]{Projection operator for sparsity constraint and the overall algorithm in Appendix \ref{app:proj_op}.}  \hfill \makecell[r]{(Page~\pageref{app:proj_op})} \\
 & \makecell[l]{Details on hidden states pruning for channel and head granularities in Appendix \ref{app:imp_details}.}  \hfill \makecell[r]{(Page~\pageref{app:imp_details})} \\
 & \makecell[l]{A reinforcement learning perspective of the proposed method in Appendix \ref{app:rl}.}  \hfill \makecell[r]{(Page~\pageref{app:rl})} \\
 & \makecell[l]{Theoretical analysis of moving average baseline for policy gradient in Appendix \ref{app:vrpg_theorical}.}  \hfill \makecell[r]{(Page~\pageref{app:vrpg_theorical})} \\
  & \makecell[l]{An in-depth discussion on the practical applicability of the proposed method in Appendix \ref{app:pratical}.}  \hfill \makecell[r]{(Page~\pageref{app:pratical})} \\
 \hline
 \hline

\multirow{9}{*}{\textbf{Results}} & \makecell[l]{\textcolor{red}{Statistics of the training time \& memory, and the inference latency in Appendix \ref{app:time_memory}.}}  \hfill \makecell[r]{\textcolor{red}{(Page~\pageref{app:time_memory})}} \\
 & \makecell[l]{\textcolor{red}{Performance after pruning and (then) finetuning in Appendix \ref{app:finetune}.}}  \hfill \makecell[r]{\textcolor{red}{(Page~\pageref{app:finetune})}} \\
 & \makecell[l]{Performance on layer pruning in Appendix \ref{app:LPruning}.}  \hfill \makecell[r]{(Page~\pageref{app:LPruning})} \\
 & \makecell[l]{Zero-shot performance on LLaMA-2-7B in Appendix \ref{app:lamma2}.}  \hfill \makecell[r]{(Page~\pageref{app:lamma2})} \\
 & \makecell[l]{Further evaluation with wider pruning rate range \ref{app:more_eval}.}  \hfill \makecell[r]{(Page~\pageref{app:more_eval})} \\
 & \makecell[l]{Comparision with approaches with weight update in Appendix \ref{app:cmp_weight_update}.}  \hfill \makecell[r]{(Page~\pageref{app:cmp_weight_update})} \\
 & \makecell[l]{Performance on Mstral-7B-Instruct-V0.3 in Appendix \ref{app:mistral}.}  \hfill \makecell[r]{(Page~\pageref{app:mistral})} \\
 & \makecell[l]{Generated samples of the pruned model in Appendix \ref{app:gen_samples}.}  \hfill \makecell[r]{(Page~\pageref{app:gen_samples})} \\
 & \makecell[l]{Random error-bar statistics in Appendix \ref{app:error_bar}.}  \hfill \makecell[r]{(Page~\pageref{app:error_bar})} \\
 \hline
 \hline

 \multirow{4}{*}{\textbf{Ablations}} & \makecell[l]{Ablations on the moving average baseline for policy gradient in Appendix \ref{app:vrpg_ablation}}  \hfill \makecell[r]{(Page~\pageref{app:vrpg_ablation})} \\
 & \makecell[l]{Ablations on projection strategy for initialization: from metric to probability in Appendix \ref{app:proj_strategy}}  \hfill \makecell[r]{(Page~\pageref{app:proj_strategy})} \\
 & \makecell[l]{More ablations with different initializations in Appendix \ref{app:diff_init}}  \hfill \makecell[r]{(Page~\pageref{app:diff_init})} \\
 & \makecell[l]{More ablations of the post-pruning modules on LLaMA-2-7B in Appendix \ref{app:post-prune}}  \hfill \makecell[r]{(Page~\pageref{app:post-prune})} \\
  & \makecell[l]{Ablations of the calibration data size in Appendix \ref{app:calib_data_size}}  \hfill \makecell[r]{(Page~\pageref{app:calib_data_size})} \\
 
\hline
\hline

\end{tabular}
\vspace{-3mm}
\caption{Summary of the Appendix materials.}
\label{tab:app_catalogs}
\vspace{-5mm}
\end{table*}

\newpage



\section{Appendix}
We discuss the following additional analyses, results, and ablations in the appendices. The catalogs are in Table \ref{tab:app_catalogs}.

\subsection{Projection Operator for Sparsity Constraint and the Overall Algorithm} \label{app:proj_op}

\textbf{Details of the Projection Operator.} In our proposed probabilistic framework, the sparsity constraint manifests itself in a feasible domain on the probability space defined in Problem \eqref{eq2-relaxed_problem}. We denote the feasible domain as $\mathcal{C} = \left \{ \mathbf{1}^{\top}\mathbf{s} \leq K \right\}  \bigcap \left \{\mathbf{s} \in [0,1]^n \right \}$. The theorem \cite{wang2013projection} below shows that the projection of a vector onto $C$ can be calculated efficiently.

\paragraph{Theorem 1.} \textit{For each vector $\mathbf{z}$ , its projection $\textbf{proj}_{\mathcal{C}}(\mathbf{z})$ in the set $C$ can be calculated as follows:} 

\begin{equation}\label{eq7-proj}
\setlength\abovedisplayskip{3pt}
\setlength\belowdisplayskip{3pt}
\begin{aligned}
\textbf{proj}_{\mathcal{C}}(\mathbf{z}) = \text{min}(1, \text{max}(0, \mathbf{z}-v_2^*\mathbf{1}))    
\end{aligned}
\end{equation}

\textit{where $v_2^* = max(0,v_1^*)$ with $v_1^*$ being the solution of the following equation}

\begin{equation}\label{eq8-solv}
\begin{aligned}
\mathbf{1}^T \left [ \text{min}(1,\text{max}(0,\mathbf{z}-v_1^*\mathbf{1})) \right ] - K = 0
\end{aligned}
\end{equation}

Equation \eqref{eq8-solv} can be solved by the bisection method efficiently. 

The theorem above as well as its proof is standard and it is a special case of the problem stated in \cite{wang2013projection}. This component, though not the highlight of our work, is included for the reader's convenience and completeness.


\textbf{Algorithm.} The pseudo-code of our overall algorithm is detailed below.
\begin{algorithm}[H]
    \caption{Pseudo-code of PG pruning}
    \label{alg:pg_pruning}
    \renewcommand{\algorithmicrequire}{\textbf{Input:}}
    \renewcommand{\algorithmicensure}{\textbf{Initialize:}}
    \begin{algorithmic}[1]
    \setlength{\baselineskip}{8pt}
        \REQUIRE target remaining ratio $r > 0$, a dense pretrained network $\mathbf{w}$, the step size $\eta > 0$, mini-batch size $B > 0$, moving average window size $T$, and calibration dataset $\mathcal{D}$
        \ENSURE Init probability $\mathbf{s}$ from any pruning metric $\mathbf{x}$, ans set moving average $\delta=0$

        \WHILE{until convergence}
            \STATE Sample a mini-batch from the entire calibration dataset: $\mathcal{D}_B=\{ (\mathbf{x}_i, \mathbf{y}_i)\}^B_{i=1} \sim \mathcal{D}$
            \STATE Sample $\mathbf{m}^{(i)}$ from $p(\mathbf{m}|s)$, $i=1,2,\dots,N_s$
            \STATE Update the moving average baseline $\delta$ via Eq. \eqref{eq:update_baseline}
            \STATE Uptate $\mathbf{s}$ via Eqs. \eqref{eq:update_s}, \eqref{eq7-proj}, and \eqref{eq8-solv}.
        \ENDWHILE
    \end{algorithmic}
\end{algorithm}

\vspace{-7mm}
\subsection{Details on Hidden States Pruning for Channel and Head Granularities} \label{app:imp_details}

\begin{figure*}
\vspace{-3mm}
    \centering
    \includegraphics[width=0.8\textwidth]{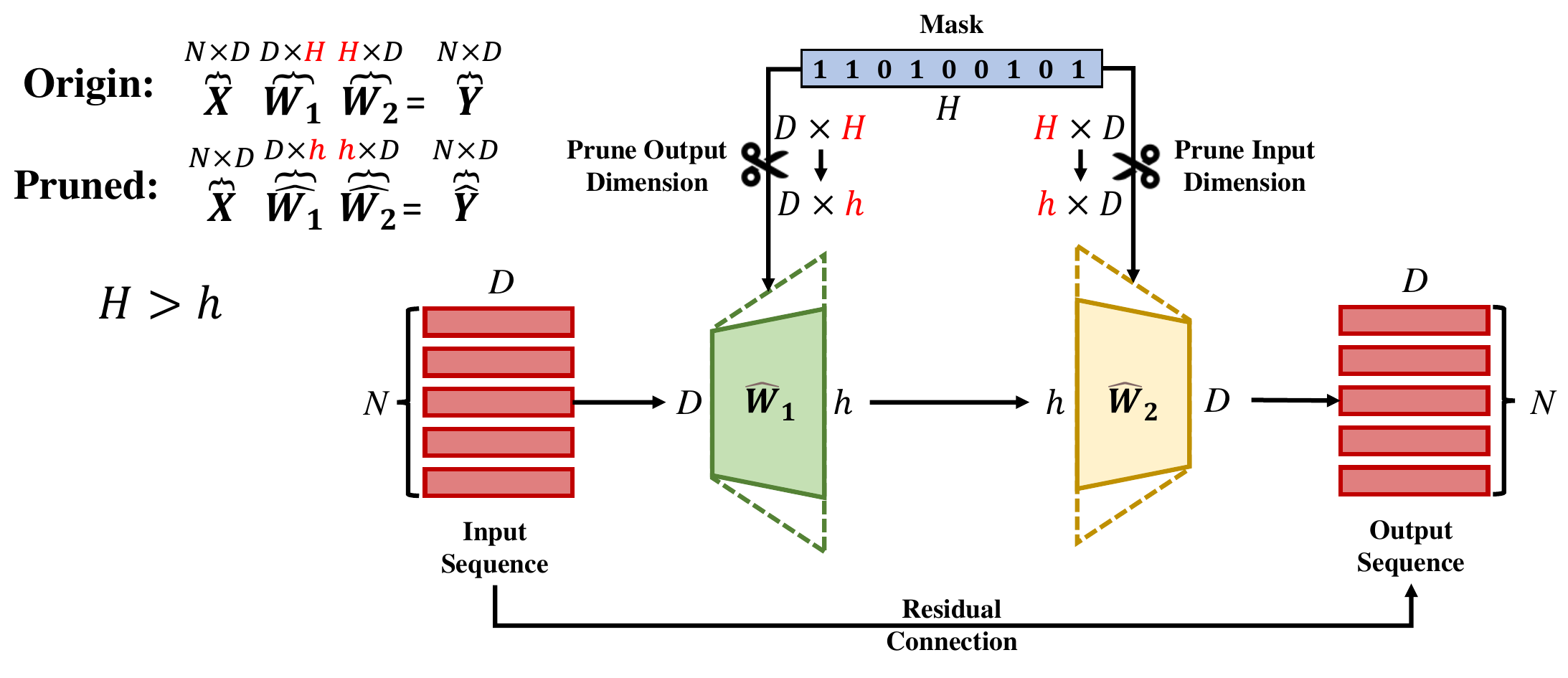}
    \vspace{-3mm}
    \caption{Output dimension is invariant for each block that might be used for residual connections, but instead prune the dimension of the intermediate hidden state.} 
    \label{fig:dim-invariant}
\vspace{-5mm}
\end{figure*}

We note that for pruning on the channel and head granularities, it must be guaranteed that the final output dimension for each block (\eg, multi-head attention, MLP) should remain, so as to facilitate the residue connections (\eg, additions) across blocks. We thus follow \cite{ma2023llm,an2024fluctuation} to prune the dimensions of the hidden states, while keeping the final output channels unchanged, ensuring that they can be added to the input through the residual connections. A conceptual figure illustrating this procedure is shown in Fig. \ref{fig:dim-invariant}.

\vspace{-2mm}
\subsection{A Reinforcement Learning Perspective} \label{app:rl}
\vspace{-1mm}

Our formulation can also be interpreted from the dense-reward model-free reinforcement learning perspective. Particularly, the heavy LLM can be viewed as the agnostic and fixed \underline{environment}. 

In terms of the Markov Decision Process (MDP) (\textbf{action} $a$, \textbf{states} $s$, \textbf{state transition probability} $p$, \textbf{reward} $r$, \textbf{discount factor} $\gamma$), the \underline{environment} takes the \textbf{action} $a$ sampled from the current \texttt{Bernoulli} \emph{policy} $\pi$ to insert the binary masks for pruning, produces the \textbf{states} $s$ as the masked/pruned network deterministically (\ie, the \textbf{state transition probability} $p$ is constantly 1), and generate the stepwise dense \textbf{reward} $r$ as the performance (\eg, the cross-entropy loss) of the pruned LLM. Since our problem exhibits dense rewards, the \textbf{discount factor} $\gamma$ is 1. 

As a result, our policy to take actions, \ie, the \texttt{Bernoulli} distribution to sample the binary masks, can be learned efficiently exploiting the \emph{policy gradient estimator} (similar to REINFORCE), without back-propagating through the agnostic and fixed \underline{environment} of the heavy LLM.

\vspace{-3mm}
\subsection{Theoretical Analysis of Moving Average Baseline for Policy Gradient} \label{app:vrpg_theorical}
\vspace{-2mm}

We give the theoretical analysis on the variance reduction technique by considering a general-purpose technique for reducing the variance of Monte Carlo method with the general problem $\mathbb{E}_{p(\mathbf{x};\mathbf{\theta})}[f(\mathbf{x})]$. We take a strategy that replacing the function $f(\mathbf{x})$ in the expectation by a substitute function $\tilde{f}(\mathbf{x})$ whose expectation $\mathbb{E}_{p(\mathbf{x};\mathbf{\theta})}[\tilde{f}(\mathbf{x})]$ is the same, but whose variance is lower. Given a function $h(\mathbf{x})$ with a known expectation $\mathbb{E}_{p(\mathbf{x};\mathbf{\theta})}[h(\mathbf{x})]$, we can easily construct such a substitute function along with the corresponding estimator as follows:
{\setlength\abovedisplayskip{3pt}
\setlength\belowdisplayskip{3pt}
\begin{align} \label{eq:prefix}
& \tilde{f}(\mathbf{x}) \!=\! f(\mathbf{x}) \!-\! \beta (h(\mathbf{x}) \!-\! \mathbb{E}_{p(\mathbf{x};\mathbf{\theta})}[h(\mathbf{x})]), \\
& \bar{\eta}_N \!=\! \frac{1}{N} \sum_{n=1}^N \tilde{f}(\hat{\mathbf{x}}^{n}) \!=\! \bar{f} \!-\! \beta (\bar{h} \!-\! \mathbb{E}_{p(\mathbf{x};\mathbf{\theta})}[h(\mathbf{x})]). \nonumber
\end{align}}

where $\hat{\mathbf{x}}^{n} \sim p(\mathbf{x};\mathbf{\theta})$ and  $\bar{f}$ and $\bar{h}$ are the sample averages. $\beta$ is a control coefficient and $h(\mathbf{x})$ is considered as  control variate.  We can show that if the variance of $h(\mathbf{x})$ is finite, the unbiasedness  the estimator Eq. \ref{eq:prefix} is maintained, \eg,
\begin{equation}
\setlength\abovedisplayskip{3pt}
\setlength\belowdisplayskip{3pt}
\begin{aligned} 
\mathbb{E}_{p(\mathbf{x};\mathbf{\theta})}[\title{f}(\mathbf{x};\beta)] = &  \mathbb{E}[\bar{f} - \beta(\bar{h} - \mathbb{E}_{h(\mathbf{x})})] \\
= & \mathbb{E}[\bar{f}] = \mathbb{E}_{p(\mathbf{x};\mathbf{\theta})}[f(\mathbf{x})]. \label{eq:unbias}
\end{aligned}
\end{equation}

For the variance of the estimator (for N = 1), we have 
{\setlength\abovedisplayskip{3pt}
\setlength\belowdisplayskip{3pt}
\begin{align}
\mathbb{V}[\tilde{f}] = & \mathbb{V}[f(\mathbf{x}) \!-\!\beta(h(\mathbf{x}) \!-\!\mathbb{E}_{p(\mathbf{x};\mathbf{\theta})}[h(\mathbf{x})])] \nonumber \\
= & \mathbb{V}[f] \!-\! 2\beta Cov[f,h] \!+\! \beta^2\mathbb{V}[h] .
\label{eq:var_est}
\end{align}}

By minimizing Eq. \ref{eq:var_est} we can find that the optimal value of the coefficient is
\begin{equation}
\setlength\abovedisplayskip{3pt}
\setlength\belowdisplayskip{3pt}
\begin{aligned}
\beta^{\ast} = \frac{\text{Cov}[f,h]}{ \mathbb{V}[h] } = \sqrt{\frac{\mathbb{V}[f]}{\mathbb{V}[h]}} \text{Corr}(f,h),
\end{aligned}\label{eq:opt_coeff}
\end{equation}

where we expressed the optimal coefficient in terms of the variance of $f$ and $h$ and the correlation coefficient $\text{Corr}(f,h)$. The effectiveness of a control variate can be measured by the variance ratio between its estimator and the original estimator: it is effective if the ratio is substantially less than 1. Using the optimal control coefficient in Eq. \ref{eq:opt_coeff}, the potential variance reduction is
{\setlength\abovedisplayskip{3pt}
\setlength\belowdisplayskip{3pt}
\begin{align}
\frac{\mathbb{V}[\tilde{f}(\mathbf{x})]}{\mathbb{V}[f(\mathbf{x})]} = & \frac{\mathbb{V}[f(\mathbf{x}-\beta(h(\mathbf{x})-\mathbb{E}_{p(\mathbf{x};\mathbf{\theta})}[h(\mathbf{x})])]}{\mathbb{V}[f(\mathbf{x})]} \nonumber \\
= & 1 - \text{Corr}(f(\mathbf{x}), h(\mathbf{x}))^2 .
\label{eq:var_red}
\end{align}}

Therefore, as long as $f(\mathbf{x})$ and $h(\mathbf{x})$ are not uncorrelated, we can always obtain a reduction in variance using control variables. In practice, the optimal $\beta^*$
will not be known and so we will usually need to estimate it empirically.

In our problem formulation of structured pruning for LLMs, $\mathbb{E}_{p(\mathbf{m}|\mathbf{s})} \mathcal{L} (\mathcal{D};\mathbf{w} \odot \mathbf{m})\nabla_{\mathbf{s}}\log(p(\mathbf{m}|\mathbf{s}))$, is a score-function gradient estimator [1], in which $p(\mathbf{m}|\mathbf{s})$ is the \texttt{Bernoulli} distribution of each module of LLMs with $\mathbf{s}$ corresponds the $\mathbf{\theta}$, $\mathbf{m}$ corresponds the $\mathbf{\theta}$ and $\mathcal{L} (\mathcal{D};\mathbf{w} \odot \mathbf{m})\nabla_{\mathbf{s}}\log(p(\mathbf{m}|\mathbf{s}))$ corresponds $f(\mathbf{x})$ in the preliminary. To reduce the variance of a score-function gradient estimator, one simple and general way is to use the score function itself as a control variate, that is $h(\mathbf{m}) = \delta \nabla_{\theta}\text{log}p(\mathbf{m}|\mathbf{s})$ and $\delta$ is an independent estimation of $\mathcal{L} (\mathcal{D};\mathbf{w} \odot \mathbf{m})$, since its expectation under the measure is zero, as 
\begin{equation}
\setlength\abovedisplayskip{3pt}
\setlength\belowdisplayskip{3pt}
\begin{aligned}
& \mathbb{E}_{p(\mathbf{m}|\mathbf{s})}[\delta \nabla_{\mathbf{s}}\text{log}p(\mathbf{m}|\mathbf{s}) ] \\
= & \delta \int p(\mathbf{m}|\mathbf{s}) \frac{\nabla_{\mathbf{s}}p(\mathbf{m}|\mathbf{s})}{p(\mathbf{m}|\mathbf{s})} d\mathbf{m} \\
= & \delta \nabla_{\mathbf{s}} \int p(\mathbf{m}|\mathbf{s}) d\mathbf{m} = \delta \nabla_{\mathbf{s}} 1 = \mathbf{0}.
\end{aligned}\label{eq:log_trick}
\end{equation}

Therefore, the estimator in Eq. \ref{eq:prefix} format is:
\begin{align}
\bar{\eta}_N & =  \frac{1}{N} \sum_{n=1}^N (\mathcal{L} (\mathcal{D};\mathbf{w} \odot \mathbf{m}^{(n)}) \label{eq:baseline} \\
& -\beta \delta)\nabla_{\mathbf{s}}\log(p(\mathbf{m}^{(n)}|\mathbf{s})); \ \mathbf{m}^{(n)} \sim p(\mathbf{m}|\mathbf{s}), \nonumber
\end{align}
where $\mathbf{m}^{(n)}$ is the sampled mask of modules. In reinforcement learning, the term $\beta \delta$ is called a baseline \cite{williams1992simple} and has historically been estimated with a running average of the cost. Note that $\delta$ needs to be estimated, we choose moving average baseline in our method, which is a commonly used baseline in policy gradient estimation \cite{zhao2011analysis,sehnke2010parameter}.

\vspace{-2mm}
\subsection{Practical Applicability Discussion} \label{app:pratical}
\vspace{-1mm}




While our optimization-based pruning method inherently requires more training time for optimization than the metric-based methods (Table \ref{tab:time_consum}), it delivers superior performance with similar memory efficiency, \ie our method is particularly well suited for scenarios where \emph{GPU memory is strictly limited}, especially when \emph{available memory only allows for inference on the unpruned model.} This case is common and practical, \emph{as SOTA models continue to grow in size, such constraints are prevalent for individual practitioners or academic institutions.}

Thus, in our main experiments, we focus on comparisons with methods that \emph{do not require weight updates and use only forward propagation.}

\begin{table}[b]
\vspace{-3mm}
    \centering
    \small
    \begin{tabular}{@{}l||c|c@{}}
    \hline
    \hline
        Method & Metric/Optim & Pruning Time \\ \hline
        LLM-Pruner & Metric & 38.11s \\ \hline
        SliceGPT & Metric & 17.15min \\ \hline
        Wanda-sp & Metric  & 36.35s \\ \hline
        Ours (Policy Gradient) & Optim & 1.56h \\ \hline
        Ours (Gumbel Softmax) & Optim & 3.47h \\ 
        \hline
        \hline
    \end{tabular}
    \vspace{-3mm}
    \caption{Pruning time comparison for LLaMA-2-7B.}
\label{tab:time_consum}
\end{table}

\textbf{Compared to Gumbel-Softmax.} Gumbel-Softmax involves backpropagation and is prone to OUT-OF-MEMORY (OOM) errors under constrained GPU memory limits. In contrast, our method avoids this issue, achieving comparable performance (Fig. \ref{fig:gumbel_perform}) while using 38\% less memory and 122\% less training time (Table \ref{tab:gumbel_time_mem}).

\textbf{Compared to metric-based pruning.} Our method requires similar memory, since it also relies only on forward propagation, but delivers significantly better performance. This is because metric-based methods depend on heuristics, whereas our approach directly optimizes the loss. Notably, increasing compute for chasing a better metric does not easily close such heuristic vs. optimization performance gap; for instance, our method outperforms and is also faster than a recent metric-based SOTA, Bonsai \cite{dery2024everybody}, across Tables \ref{tab:channel_ppl_results}, \ref{tab:zero_shot_llama3}, \ref{tab:wtf_zero_shot_llama2}, \ref{tab:wtf_zero_shot_llama3}, \ref{tab:zero_shot_llama2}, \ref{tab:lwo_pr_llama3}.

\textbf{Compared to weight-update methods with similar training time.} These approaches also require back-propagation and thus suffer from OOM issues under memory constraints. In contrast, once our efficient pruning is applied, the resulted smaller model enables feasible finetuning. We show that such pruning-then-finetuning further enhances performance in memory-constrained settings (Tables \ref{tab:wtf_zero_shot_llama2}, \ref{tab:wtf_zero_shot_llama3}, and \ref{tab:cmp_weight_update}). Especially, Table \ref{tab:cmp_weight_update} demonstrates that our method, when followed by fine-tuning, outperforms SOTA weight-update methods such as FLAP \cite{an2024fluctuation} and Search-LLM \cite{shen2024search}.

\vspace{-2mm}
\subsection{Statistics of the Training Time \& Memory, and the Inference Latency} \label{app:time_memory}
\vspace{-3mm}

\begin{table}[ht]
\centering
\small
\begin{tabular}{@{}l||c|c||c|c@{}}
\hline
\hline

\multirow{2}{*}{Method} & \multicolumn{2}{c||}{7B} & \multicolumn{2}{c}{13B} \\
\cline{2-5}
& Min & Max & Min & Max \\
\hline

Wanda-sp & 17.5 & 20.3 & 29.5 & 36.9 \\
\hline
Ours & 17.2 & 17.4 & 34.1 & 35.8 \\

\hline
\hline
\end{tabular}
\vspace{-3mm}
\caption{Memory requirements (GB) for channel and head pruning on LLaMA-2-7B/13B.}
\label{tab:mem_req}
\vspace{-4mm}
\end{table}

\begin{table}[ht]
\vspace{-2mm}
\centering
\small
\setlength{\tabcolsep}{1.2mm} {
\begin{tabular}{@{}l||c||c|c|c|c@{}}
\hline
\hline
Method & P.R & \#Params & Memory & Latency & PPL \\
\hline

LLM-Pruner & \multirow{3}{*}{30$\%$} & 4.837 & \textbf{9290.54} & 53.53 & 27.13 \\
SliceGPT &                           & 5.293 & 10181.81 & 50.24 & 22.29 \\
Ours &                               & \textbf{4.796} & 9338.24 & \textbf{46.94} & \textbf{12.68} \\
\hline

LLM-Pruner & \multirow{3}{*}{40$\%$} & 4.197 & \textbf{8069.55} & \textbf{36.75} & 53.21 \\
SliceGPT &                           & 4.501 & 8826.01 & 46.84 & 39.21  \\
Ours &                               & \textbf{4.149} & 8096.25 & 42.85 & \textbf{15.95}  \\
\hline

LLM-Pruner & \multirow{3}{*}{50$\%$} & 3.539 & \textbf{6815.05} & \textbf{31.49} & 171.57  \\
SliceGPT &                           & 3.730 & 7274.01 & 41.73 & 65.92 \\
Ours &                               & \textbf{3.500} & 6880.92 & 34.62 & \textbf{27.63} \\
\hline

\hline
\hline
\end{tabular}
}
\vspace{-3mm}
\caption{\#Params (B), memory requirements (MiB), latency (s) and WikiText2 perplexity (\ie PPL) of LLaMA-2-7B. Experiments are conducted on NVIDIA A100 40G, with 2048 sequence length and 4 batch size for full GPU utilization. P.R. is short for pruning rate.}
\label{tab:latency}
\vspace{-4mm}
\end{table}

\begin{table*}[ht]
\centering
\small
\begin{tabular}{@{}l||c||c||c|c|c|c|c||c@{}}
\hline
\hline
Method & PruneRate & PPL $\downarrow$ & PIQA & HellaSwag & WinoGrande & ARC-e & ARC-c & Average \\
\hline

Dense & 0$\%$ & 12.19 & 78.02 & 57.17 & 68.43 & 76.30 & 43.51 & 64.69 \\
\hline

LLM-Pruner & \multirow{5}{*}{30$\%$} & 33.45 & 74.10 & 46.61 & 58.17 & 64.31 & 33.62 & 55.36 \\
SliceGPT &                           & 78.59 & 74.70 & \textbf{64.29} & 61.96 & 57.49 & 36.69 & 59.03 \\
Bonsai &                             & 33.23 & 75.03 & 49.69 & 62.19 & 67.34 & 32.25 & 57.30 \\
Wanda-sp &                           & 32.01 & 73.88 & 50.08 & 62.19 & 67.09 & 34.47 & 57.54 \\
Ours &                               & \textbf{25.34} & \textbf{76.01} & 51.80 & \textbf{64.33} & \textbf{67.93} & \textbf{36.86} & \textbf{59.39} \\
\hline

LLM-Pruner & \multirow{5}{*}{40$\%$} & 40.21 & 70.29 & 40.45 & 53.04 & 53.03 & 27.30 & 48.82 \\
SliceGPT &                           & 175.67 & 65.29 & \textbf{56.77} & \textbf{60.06} & 42.68 & \textbf{31.74} & 51.31 \\
Bonsai &                             & 44.71 & 72.36 & 45.10 & 58.80 & 59.64 & 30.03 & 53.19 \\
Wanda-sp &                           & 43.71 & 70.40 & 42.73 & 52.72 & 57.24 & 29.95 & 50.61 \\
Ours &                               & \textbf{29.43} & \textbf{72.74} & 45.75 & 55.72 & \textbf{61.36} & 31.06 & \textbf{53.33} \\
\hline

LLM-Pruner & \multirow{5}{*}{50$\%$} & 44.83 & \textbf{67.30} & 35.47 & 51.93 & 48.23 & 21.84 & 44.95 \\
SliceGPT &                           & 296.97 & 58.65 & \textbf{46.83} & \textbf{55.09} & 36.99 & \textbf{28.33} & 45.18 \\
Bonsai &                             & 62.95 & 66.70 & 40.16 & 54.30 & 49.83 & 26.53 & \textbf{47.50} \\
Wanda-sp &                           & 110.12 & 63.27 & 32.71 & 52.72 & 43.48 & 20.73 & 42.58 \\
Ours &                               & \textbf{39.46} & 67.03 & 36.42 & 52.41 & \textbf{50.17} & 24.15 & 46.04 \\

\hline
\hline
\end{tabular}
\vspace{-3mm}
\caption{Perplexity (PPL) and Accuracies ($\%$) of LLaMA-2-7B for 5 zero-shot tasks with pruning rates from 30$\%$ to 50$\%$ after weight fine-tuning on Alapca dataset.}
\label{tab:wtf_zero_shot_llama2}
\vspace{-6mm}
\end{table*}

Our training times for channel and head pruning on LLaMA-2-7B and LLaMA-2-13B are 1.76 and 2.72 hours, respectively. The statistics of the time consumption of different pruning methods on LLaMA-2-7B are shown in Table \ref{tab:time_consum}. Although our method is slower than metric-based methods such as Wanda-sp \cite{an2024fluctuation}, the trade-off is justified by the substantial performance gains delivered by our optimization-based approach.


The GPU memory requirements for channel and head pruning on LLaMA-2-7B and LLaMA-2-13B for our method, as well as the representative metric-based method, \eg, Wanda-sp, are illustrated in Table \ref{tab:mem_req}. We do not compare it to LLM-Pruner and SliceGPT because 1) the LLM-Pruner requires much more memory for back-propagation (therefore the authors also used the CPU memory), 2) the original implementation of SliceGPT also used both CPU and GPU memory for computations. Table \ref{tab:mem_req} shows that our method exhibits a similar GPU memory requirement to the efficient Wanda-sp, as we only need the forward pass of the LLM. The slight additional memory required by our method comes from the need to store the \texttt{Bernoulli} parameters $\mathbf{s}$ and sampled masks $\mathbf{m}$.

\begin{table*}[ht]
\centering
\small
\begin{tabular}{@{}l||c||c||c|c|c|c|c||c@{}}
\hline
\hline
Method & PruneRate & PPL $\downarrow$ & PIQA & HellaSwag & WinoGrande & ARC-e & ARC-c & Average \\
\hline

Dense & 0$\%$ & 14.14 & 79.71 & 60.19 & 72.61 & 80.09 & 50.34 & 68.59 \\
\hline

LLM-Pruner & \multirow{5}{*}{30$\%$} & 35.11 & \textbf{74.64} & 46.93 & 60.22 & \textbf{66.16} & 34.13 & 56.42 \\
SliceGPT &                           & 226.39 & 70.29 & \textbf{56.47} & 60.06 & 53.20 & 34.81 & 54.97 \\
Bonsai &                             & 42.59 & 71.87 & 45.17 & 59.51 & 66.50 & \textbf{36.52} & 55.91 \\
Wanda-sp &                           & 38.04 & 70.84 & 44.11 & 59.43 & 62.96 & 34.04 & 54.28 \\
Ours &                               & \textbf{33.91} & 74.48 & 46.62 & \textbf{63.69} & 65.70 & 34.30 & \textbf{56.96} \\
\hline

LLM-Pruner & \multirow{5}{*}{40$\%$} & 47.83 & \textbf{71.54} & 40.71 & 55.40 & \textbf{62.16} & 28.92 & 51.75 \\
SliceGPT &                           & 523.05 & 63.66 & \textbf{42.75} & 53.12 & 41.88 & 27.65 & 45.81 \\
Bonsai &                             & 57.31 & 69.58 & 39.47 & 53.98 & 57.24 & 28.67 & 49.79 \\
Wanda-sp &                           & 56.32 & 65.18 & 36.33 & 54.77 & 51.56 & 24.32 & 46.43 \\
Ours &                               & \textbf{47.28} & 70.56 & 41.09 & \textbf{59.98} & 59.97 & \textbf{29.01} & \textbf{52.12} \\
\hline

LLM-Pruner & \multirow{5}{*}{50$\%$} & 68.14 & \textbf{67.95} & 35.81 & 53.12 & \textbf{53.91} & 26.36 & 47.43 \\
SliceGPT &                           & 963.42 & 60.83 & \textbf{37.04} & 52.25 & 37.21 & 25.26 & 42.52 \\
Bonsai &                             & 88.72 & 62.89 & 34.84 & 52.80 & 47.73 & 24.15 & 44.48 \\
Wanda-sp &                           & 84.53 & 61.42 & 32.12 & 52.72 & 41.83 & 21.07 & 41.83 \\
Ours &                               & \textbf{67.48} & 67.08 & 35.84 & \textbf{54.38} & 53.54 & \textbf{26.45} & \textbf{47.46} \\

\hline
\hline
\end{tabular}
\vspace{-3mm}
\caption{Perplexity (PPL) and Accuracies ($\%$) of LLaMA-3-8B for 5 zero-shot tasks with pruning rates from 30$\%$ to 50$\%$ after weight fine-tuning on Alapca dataset.}
\label{tab:wtf_zero_shot_llama3}
\vspace{-2mm}
\end{table*}

\begin{table*}[t]
\centering
\small
\begin{tabular}{@{}l||c||c|c||c|c||c||c|c@{}}
\hline
\hline
    \multirow{2}{*}{Method} & \multirow{2}{*}{PruneRate} & \multicolumn{2}{c||}{LLaMA} & \multicolumn{2}{c||}{LLaMA-2} & LLaMA-3 & \multicolumn{2}{c}{Vicuna} \\
    \cline{3-9}
    & & 7B & 13B & 7B & 13B & 8B & 7B & 13B \\
    \hline
    Dense  & 0$\%$ & 12.62 & 10.81 & 12.19 & 10.98 & 14.14 & 16.24 & 13.50 \\
    \hline

    Layerwise-PPL & \multirow{3}{*}{30$\%$} & 31.65 & 24.23 & 24.83 & 20.52 & 45.47 & 37.99 & 29.85 \\
    SLEB &  & 27.36 & \textbf{20.45} & 23.43 & \textbf{19.97} & 37.92 & 29.40 & 26.37 \\
    Ours &  & \textbf{24.45} & 24.44 & \textbf{23.20} & 21.93 & \textbf{36.42} & \textbf{29.16} & \textbf{24.68} \\
    \hline


    Layerwise-PPL & \multirow{3}{*}{40$\%$} & 54.97 & 50.57 & 41.45 & 32.48 & 75.12 & 64.96 & 54.12 \\
    SLEB &  & 44.65 & \textbf{32.79} & 40.26 & \textbf{30.16} & 73.61 & \textbf{48.99} & 43.12 \\
    Ours &  & \textbf{42.73} & 39.07 & \textbf{38.26} & 30.99 & \textbf{63.70} & 54.37 & \textbf{35.73} \\
    \hline


    Layerwise-PPL & \multirow{3}{*}{50$\%$} & 107.12 & 183.93 & 126.08 & 78.04 & 393.18 & 517.46 & 153.53 \\
    SLEB &  & 108.87 & 77.38 & 131.49 & \textbf{55.23} & 303.03 & 146.12 & 92.32 \\
    Ours &  & \textbf{94.97} & \textbf{66.38} & \textbf{104.37} & 69.92 & \textbf{295.39} & \textbf{126.24} & \textbf{84.90} \\
    
\hline
\hline
\end{tabular}
\vspace{-3mm}
\caption{Results on \emph{layers} pruning. Our method is initialized by Layerwise-PPL (please also refer to Appendix \ref{app:diff_init} for detailed discussion about initializations). All the methods are calibrated using the C4 dataset and validated on the WikiText2 dataset \wrt perplexity.}
\label{tab:layer_ppl_results}
\vspace{-6mm}
\end{table*}

We note that for the same pruning rate (\ie, similar remaining \#Params), the inference latencies of pruned models from different structural pruning methods are expected to be comparable, as the inference latency is mainly affected by the \#Params given the same architecture. We validate this in Table \ref{tab:latency}. Table \ref{tab:latency} demonstrates that, given the same pruning rates, our pruned model has very much close \#Params, memory, and inference latencies to that pruned by LLM-Pruner, while our perplexity significantly outperformed all the counterparts. We note that under the same pruning rates, SliceGPT often possesses different (higher) \#Params, memory, and inference latencies than our method and LLM-Pruner, potentially because SliceGPT alters the network structure during the pruning.






\vspace{-2mm}
\subsection{Performance after Pruning and (Then) Finetuning} \label{app:finetune}
\vspace{-1mm}

We note that after pruning, it becomes affordable to finetune a smaller pruned model. Therefore, following the idea from \cite{ma2023llm}, we finetune the post-pruning model \wrt the perplexity with LoRA \cite{hu2021lora}. Specifically, we utilize 4k samples from the Alpaca \cite{taori2023stanford} dataset, which has a sequence length of 1024. For all weight fine-tuning experiments, we use $\textit{lora\_r} \!=\! 16$, $\textit{lora\_alpha} \!=\! 10$, and use default values for all other hyperparameters in the HuggingFace PEFT package \cite{mangrulkar2022peft}.

The cross-dataset performance on WikiText of the post-pruning fine-tuned model for LLaMA-2-7B and LLaMA-3-8B is illustrated in Tables \ref{tab:wtf_zero_shot_llama2} and \ref{tab:wtf_zero_shot_llama3}, which demonstrate that our method achieves consistently superior performance before and after fine-tuning. Compared with the pre-finetuned model, the performance of most post-finetuned models shows significant improvements, and our models remain the best for most cases after finetuning, which validates our potential for narrowing the performance gap after pruning and for being applicable in practical use.

\vspace{-2mm}
\subsection{Results on Layer Pruning} \label{app:LPruning}
\vspace{-1mm}

We also validate the layer granularity by pruning the entire transformer layer, consisting of an MHA module and a MLP module. Note that pruning on the layer granularity is less exploited for LLMs, thus in this experiment, we use the lightweight Layerwise-PPL \cite{kim2024shortened} for initialization, and compare our method with Layerwise-PPL \cite{kim2024shortened} and SLEB \cite{song2024sleb}.

We illustrate the results on layer pruning in Table \ref{tab:layer_ppl_results}, which show that our method generally achieves better performance than the baseline methods, especially at pruning rates above 40\%. For LLaMA-13B and LLaMA-2-13B with a moderate pruning rate of 30\%, our method performs comparably to Layerwise-PPL. This suggests that with coarser layer granularity, the search space may be limited, and larger 13B models with more redundancy benefit from metric-based pruning at lower rates.

\begin{table*}[h]
\centering
\small
\begin{tabular}{@{}l||c||c||c|c|c|c|c||c@{}}
\hline
\hline
Method & PruneRate & PPL $\downarrow$ & PIQA & HellaSwag & WinoGrande & ARC-e & ARC-c & Average \\
\hline

Dense & 0$\%$ & 12.19 & 78.02 & 57.17 & 68.43 & 76.30 & 43.51 & 64.69 \\
\hline

LLM-Pruner & \multirow{5}{*}{30$\%$} & 38.94 & 71.81 & 43.64 & 54.06 & 63.42 & 30.30 & 52.64 \\
SliceGPT &                           & 40.40 & 72.31 & \textbf{60.11} & \textbf{63.22} & 53.10 & 32.00 & 56.15 \\
Bonsai &                             & 39.01 & 73.94 & 47.05 & 60.93 & 59.93 & 30.37 & 54.44 \\
Wanda-sp &                           & 49.13 & 71.60 & 46.62 & 60.30 & 63.01 & 34.04 & 55.11 \\
Ours &                               & \textbf{28.18} & \textbf{75.41} & 50.34 & 61.60 & \textbf{66.03} & \textbf{35.58} & \textbf{57.79} \\
\hline

LLM-Pruner & \multirow{5}{*}{40$\%$} & 68.48 & 67.52 & 35.76 & 51.70 & 48.31 & 24.65 & 45.59 \\
SliceGPT &                           & 73.76 & 65.40 & \textbf{48.91} & \textbf{60.38} & 42.13 & 26.88 & 48.74 \\
Bonsai &                             & 69.18 & 68.44 & 40.63 & 55.41 & 48.11 & 26.19 & 47.75 \\
Wanda-sp &                           & 78.45 & 64.63 & 35.61 & 52.17 & 48.11 & 25.51 & 45.21 \\
Ours &                               & \textbf{39.81} & \textbf{71.11} & 42.44 & 55.72 & \textbf{56.94} & \textbf{28.50} & \textbf{50.94} \\
\hline

LLM-Pruner & \multirow{5}{*}{50$\%$} & 190.56 & 59.52 & 29.74 & 50.11 & 36.48 & 21.84 & 39.54 \\
SliceGPT &                           & 136.33 & 59.47 & \textbf{37.96} & \textbf{56.27} & 33.63 & \textbf{22.78} & \textbf{42.02} \\
Bonsai &                             & 216.85 & 59.52 & 32.63 & 53.12 & 33.54 & 22.61 & 40.28 \\
Wanda-sp &                           & 206.94 & 54.30 & 26.81 & 52.80 & 29.12 & 19.20 & 36.45 \\
Ours &                               & \textbf{65.21} & \textbf{61.80} & 30.94 & 52.64 & \textbf{40.11} & 20.47 & 41.19 \\

\hline
\hline
\end{tabular}
\vspace{-3mm}
\caption{Perplexity (PPL) and accuracies ($\%$) of LLaMA-2-7B for 5 zero-shot tasks with 30\% - 50\% pruning rates.}
\label{tab:zero_shot_llama2}
\vspace{-2.5mm}
\end{table*}

\vspace{-2.5mm}
\subsection{Zero-shot Performance on LLaMA-2-7B} \label{app:lamma2}
\vspace{-1mm}

We validate the zero-shot performance of LLaMA-2-7B with pruning rates from 30\% to 50\%, shown in Table \ref{tab:zero_shot_llama2}. We note that the overall performance is in general superior to the baselines, though using only the C4 dataset for pruning might introduce a negative influence on some particular cross-dataset zero-shot tasks such as WinoGrande \cite{sakaguchi2021winogrande} and Hellaswag \cite{zellers2019hellaswag}. 

\vspace{-2mm}
\subsection{Further Evaluation on Wider Pruning Rate Range} \label{app:more_eval}
\vspace{-2mm}

For a more comprehensive validation of the proposed method, we experiment on LLaMA-3-8B with a wider range of pruning rate, from 10\% to 50\%, following the same settings of the main results. Beyond WikiText2 perplexity and 5 zero-shot tasks, the comparison on MMLU benchmark \cite{hendrycks2020measuring} for five-shot and additional zero-shot task, LAMBADA \cite{paperno2016lambada}, RACE \cite{lai2017race} and BoolQ \cite{clark2019boolq}, are also included. The results shown in Table \ref{tab:lwo_pr_llama3} demonstrate the consistent superiority of our method across a wide range of sparsity levels.

\begin{table*}[h]
\centering
\tiny
\begin{tabular}{@{}l||c||c||c|c|c||c|c|c|c|c||c@{}}
\hline
\hline
Method & PruneRate                   & PPL $\downarrow$ & LAMBADA & RACE & BoolQ & PIQA & HellaSwag & WinoGrande & ARC-e & ARC-c & MMLU \\
\hline

Dense & 0\%                          & 14.14 & 69.14 & 40.29 & 81.41 & 79.71 & 60.19 & 72.61 & 80.09 & 50.34 & 66.58  \\
\hline


LLM-Pruner & \multirow{5}{*}{10$\%$} & 19.25 & 53.85 & 37.32 & 73.24 & 77.04 & 52.93 & 68.11 & 73.44 & 39.50 & 48.37 \\ 
SliceGPT &  & 39.14 & 59.67 & \textbf{40.29} & \textbf{80.58} & 75.57 & 54.78 & 68.35 & 72.56 & 40.87 & 55.38 \\ 
Bonsai &  & 20.43 & 54.12 & 38.75 & 75.69 & 77.64 & 54.96 & 70.32 & 75.92 & \textbf{43.00} & 39.56 \\ 
Wanda-sp &  & 35.94 & 28.58 & 32.25 & 57.12 & 69.64 & 42.86 & 64.64 & 65.07 & 32.68 & 29.00 \\ 
Ours &  & \textbf{18.73} & \textbf{62.63} & 39.62 & 79.39 & \textbf{78.94} & \textbf{56.88} & \textbf{69.45} & \textbf{76.18} & 41.55 & \textbf{56.38} \\
\hline

LLM-Pruner & \multirow{5}{*}{20$\%$} & 28.62 & 37.45 & 32.63 & 55.96 & 74.92 & 42.94 & 59.19 & 65.57 & 32.51 & 26.78 \\
SliceGPT &  & 84.99 & 46.52 & \textbf{39.52} & \textbf{76.12} & 73.23 & 48.24 & 63.69 & 64.77 & 34.13 & 33.55 \\
Bonsai &  & 29.05 & 48.36 & 35.41 & 64.16 & 75.46 & 47.00 & 67.01 & 65.61 & 35.41 & 29.60 \\
Wanda-sp &  & 47.43 & 22.76 & 31.10 & 53.21 & 67.90 & 39.27 & 60.38 & 58.50 & 29.01 & 27.96 \\
Ours &  & \textbf{26.92} & \textbf{51.02} & 37.03 & 74.22 & \textbf{76.28} & \textbf{51.00} & \textbf{67.64} & \textbf{69.15} & \textbf{35.41} & \textbf{44.99} \\
\hline 

LLM-Pruner & \multirow{5}{*}{30$\%$} & 40.18 & 28.74 & 30.63 & 58.56 & 71.38 & 37.84 & 55.64 & 57.78 & 27.21 & 25.36 \\ 
SliceGPT &  & 183.94 & 29.17 & \textbf{36.75} & \textbf{68.20} & 68.34 & \textbf{53.92} & 57.22 & 49.41 & 28.07 & 25.89 \\ 
Bonsai &  & 80.89 & 15.50 & 31.29 & 45.29 & 64.53 & 36.10 & 55.09 & 47.64 & 22.52 & 23.41 \\
Wanda-sp &  & 92.14 & 13.87 & 28.52 & 51.28 & 59.74 & 31.46 & 52.64 & 44.02 & 19.88 & 26.25 \\
Ours &  & \textbf{38.99} & \textbf{44.81} & 35.41 & 66.15 & \textbf{72.25} & 43.56 & \textbf{59.04} & \textbf{59.85} & \textbf{29.44} & \textbf{27.38} \\
\hline

LLM-Pruner & \multirow{5}{*}{40$\%$} & 70.60 & 14.09 & 28.13 & 59.57 & 66.26 & 31.90 & 54.06 & 49.74 & 22.52 & 25.36 \\
SliceGPT &  & 354.24 & 16.28 & \textbf{33.4} & \textbf{62.87} & 61.53 & \textbf{39.98} & 52.80 & 36.66 & \textbf{25.17} & 26.10 \\
Bonsai &  & 204.61 & 2.04 & 23.35 & 46.27 & 58.81 & 29.43 & 48.93 & 33.21 & 18.15 & 25.09 \\
Wanda-sp &  & 213.47 & 8.73 & 28.23 & 52.78 & 56.58 & 27.46 & 50.35 & 32.07 & 17.06 & 25.57 \\
Ours &  &\textbf{ 63.85} & \textbf{30.80} & 32.63 & 61.96 & \textbf{67.63} & 37.36 & \textbf{56.91} & \textbf{50.67} & 24.91 & \textbf{27.50} \\
\hline

LLM-Pruner & \multirow{5}{*}{50$\%$} & 145.66 & 4.37 & 24.50 & 45.53 & 61.15 & 29.10 & \textbf{51.93} & 39.98 & 19.36 & 24.36 \\
SliceGPT &  & 841.20 & 7.99 & \textbf{30.72} & 57.00 & 56.37 & \textbf{32.66} & 48.38 & 32.45 & \textbf{22.10} & 24.16 \\
Bonsai &  & 440.86 & 0.25 & 22.10 & 42.20 & 55.66 & 26.94 & 50.51 & 30.64 & 17.83 & 24.35 \\
Wanda-sp &  & 413.86 & 3.07 & 23.25 & 45.99 & 55.39 & 27.07 & 49.72 & 29.59 & 18.26 & 24.73 \\
Ours &  & \textbf{119.75} & \textbf{17.43} & 26.79 & \textbf{61.80} & \textbf{62.51} & 30.89 & 51.85 & \textbf{41.12} & 20.65 & \textbf{25.33} \\ 

\hline
\hline
\end{tabular}
\vspace{-3mm}
\caption{Perplexity (PPL) and accuracies ($\%$) of LLaMA-3-8B for 8 zero-shot tasks and MMLU benchmark in five-shot with pruning rates from 10$\%$ to 50$\%$.}
\label{tab:lwo_pr_llama3}
\vspace{-6mm}
\end{table*}

\vspace{-3mm}
\subsection{Comparision with Approaches with Weight Update} \label{app:cmp_weight_update}
\vspace{-2mm}

We additionally conduct experiments of performance comparison with approaches that involve weight update \cite{shen2024search, an2024fluctuation}. We follow the pruning settings of \cite{shen2024search}, in which we calibrate on WikiText2 dataset and evaluate perplexity on it with a sequence length of 2048. For zero-shot tasks evaluation, we follow the procedure applied in \cite{shen2024search}. We compared our vanilla method (pruning only, without weight update) with \cite{shen2024search, an2024fluctuation}, denoted as ours (prune-only) in Table \ref{tab:cmp_weight_update}. The experiments are performed on LLaMA-7B consistent with \cite{shen2024search}. 

Additionally, since fine-tuning becomes feasible after pruning smaller models, we also included results for our prune-then-finetune approach for comparison. The results demonstrate that our pruning-only method achieves comparable performance to \cite{shen2024search}, while the prune-then-finetune approach, involving weight update, outperforms \cite{shen2024search} in the majority of cases.

\begin{table*}[h]
\centering
\small
\begin{tabular}{@{}l||c||c||c|c|c|c|c||c@{}}
\hline
\hline
Method & PruneRate                   & PPL $\downarrow$ & PIQA & HellaSwag & WinoGrande & ARC-e & ARC-c & Average \\
\hline

Dense & 0\%                       & 5.68 & 78.35 & 72.99 & 67.01 & 67.45 & 41.38 & 65.44 \\
\hline

FLAP  &  \multirow{4}{*}{10$\%$}  & 6.34 & 75.41 & 68.68 & 67.01 & 65.78 & 38.48 & 63.07 \\
search-llm &                      & \textbf{6.10} & 76.88 & 70.71 & \textbf{67.56} & 68.39 & 40.10 & 64.73 \\
ours (prune-only) &               & 6.17 & 77.53 & \textbf{71.85} & 66.14 & 69.23 & 40.87 & 65.12 \\
ours (prune-then-finetune) &      & 7.03 & \textbf{77.64} & 71.53 & 67.32 & \textbf{69.49} & \textbf{41.98} & \textbf{65.59} \\
\hline

FLAP  &  \multirow{4}{*}{20$\%$}  & 7.40 & 74.21 & 64.98 & 64.40 & 59.89 & 37.80 & 60.26 \\
search-llm &                      & 6.89 & 74.92 & 67.29 & \textbf{64.64} & 64.23 & 36.52 & 61.52 \\
ours (prune-only) &               & 7.07 & 74.92 & \textbf{68.32} & 61.56 & 62.63 & 37.20 & 60.93 \\
ours (prune-then-finetune) &      & \textbf{6.29} & \textbf{76.11} & 68.19 & 63.38 & \textbf{66.24} & \textbf{38.99} & \textbf{62.58} \\

\hline
\hline
\end{tabular}
\vspace{-3mm}
\caption{Perplexity (PPL) and accuracies ($\%$) of LLaMA-7B for 5 zero-shot tasks with pruning rates from 10$\%$ to 20$\%$, compared with approaches with weight update, FLAP \cite{an2024fluctuation} and search-llm \cite{shen2024search}.}
\label{tab:cmp_weight_update}
\vspace{-2mm}
\end{table*}

\begin{table*}[h!]
\centering
\small
\begin{tabular}{@{}l||c||c||c|c|c|c|c||c@{}}
\hline
\hline
Method & PruneRate & PPL $\downarrow$ & PIQA & HellaSwag & WinoGrande & ARC-e & ARC-c & Average \\
\hline

Dense & 0$\%$ & 12.70 & 81.77 & 64.84 & 74.51 & 84.22 & 57.34 & 72.54 \\
\hline

LLM-Pruner & \multirow{3}{*}{30$\%$} & \textbf{30.32} & 69.58 & 41.52 & 57.77 & 53.99 & 28.58 & 50.29 \\
Wanda-sp &                               & 47.30 & 75.68 & 49.94 & 62.35 & 64.90 & 36.60 & 57.89 \\
Ours &                               & 31.87 & \textbf{76.49} & \textbf{52.69} & \textbf{64.48} & \textbf{67.76} & \textbf{36.77} & \textbf{59.64} \\
\hline

LLM-Pruner & \multirow{3}{*}{40$\%$} & 49.30 & 65.18 & 34.79 & 52.80 & 46.42 & 23.89 & 44.62 \\
Wanda-sp &                               & 76.45 & 68.01 & 38.75 & 52.64 & 52.36 & 26.28 & 47.61 \\
Ours &                               & \textbf{43.02} & \textbf{68.61} & \textbf{40.80} & \textbf{56.67} & \textbf{54.80} & \textbf{27.82} & \textbf{49.74} \\
\hline

LLM-Pruner & \multirow{3}{*}{50$\%$} & 86.24 & 61.31 & 30.64 & 49.64 & 37.67 & 22.52 & 40.36 \\
Wanda-sp &                               & 407.33 & 56.69 & 29.08 & 49.25 & 32.36 & 21.59 & 37.79 \\
Ours &                               & \textbf{74.25} & \textbf{65.18} & \textbf{35.02} & \textbf{51.06} & \textbf{48.15} & \textbf{22.61} & \textbf{44.40} \\

\hline
\hline
\end{tabular}
\vspace{-3mm}
\caption{Perplexity (PPL) and accuracies ($\%$) of Mistral-7B-Instruct-v0.3 for 5 zero-shot tasks with 30\% - 50\% pruning rates.}
\label{tab:mistral}
\vspace{-6mm}
\end{table*}

\vspace{-2mm}
\subsection{Results on Mistral-7B-Instruct-v0.3} \label{app:mistral}
\vspace{-2mm}

To further validate the performance of the proposed method on more LLMs, we additionally perform experiments on Mistral-7B-Instruct-v0.3 \cite{jiang2023mistral}, which calibrates on C4 dataset and evaluates on the WikiText2 dataset (\eg, cross-dataset setting, as those in our Table \ref{tab:channel_ppl_results}). We note that the original implementations of SliceGPT \cite{ashkboos2024slicegpt} and Bonsai \cite{dery2024everybody} were based on LLaMA-2, which do not trivially adapt to the Mistral model directly, therefore, we exclude SliceGPT and Bonsai for comparison. 

The results, including both perplexity and the zero-shot performance, on Mistral-7B-Instruct-v0.3 in Table \ref{tab:mistral} demonstrate the consistent superiority of our method across various LLMs.

\vspace{-3mm}
\subsection{Generated Samples of the Pruned Model} \label{app:gen_samples}
\vspace{-1mm}

We provide some generated sentences of the pruned models. Table \ref{tab:gen_samples} illustrates the generated sentences of LLaMA-2-7B with the pruning rate of 30\%, from different pruning methods, where the input prompts are adopted from \cite{ma2023llm}. We observe that the generated content from our method not only maintains superior coherence and innovation but also is more factual and professional despite a high pruning rate (30\%). It demonstrates that our method optimizes the balance between knowledge retention and performance in the compression process, ensuring the quality and diversity of the generated text. 

\vspace{-3mm}
\subsection{Random Error-Bar Statistic} \label{app:error_bar}

The standard deviation statistics of our method are shown in Table \ref{tab:standard_deviation}. Theoretically, the variance arises from stochastic sampling from \texttt{Bernoulli} distribution in the policy gradient optimization if the initialization is fixed. Thus, we fixed initialization as Wanda-sp to calculate the standard deviation of the proposed method. Experiments of head and channel pruning, along with layer pruning, are executed using LLaMA-2-7B for 10 run trials, demonstrating reasonable deviation.
\begin{table}[hb]
\vspace{-2mm}
    \centering
    \small
    \setlength{\tabcolsep}{1.5mm}{
    \begin{tabular}{@{}l||c|c|c@{}}
    \hline
    \hline
    \multirow{2}{*}{Granularity} & \multicolumn{3}{c}{PruneRate} \\
    \cline{2-4}
     & 30\% & 40\% & 50\% \\
    \hline
    Head \& Channel & 28.18$\pm$\scriptsize{1.83} & 39.81$\pm$\scriptsize{1.41} & 65.21$\pm$\scriptsize{2.52} \\
    \hline
    Layer & 23.20$\pm$\scriptsize{0.67} & 38.26$\pm$\scriptsize{2.68} & 104.37$\pm$\scriptsize{1.05} \\
    \hline
    \hline
    \end{tabular}}
    \vspace{-3mm}
    \caption{Mean and standard deviation of our method for LLaMA-2-7B.}
    \label{tab:standard_deviation}
\vspace{-6mm}
\end{table}

\subsection{Ablations on the Moving Average Baseline for Policy Gradient} \label{app:vrpg_ablation}

We conduct experiments on pruning channels and heads of LLaMA-2-7B/13B with/without the \emph{Moving Average Baseline} in policy gradient. Table \ref{tab:wo_mab} illustrates the effectiveness of the moving average baseline in the policy gradient estimator for our proposed pruning method.

Moreover, we also tested all the hyper-parameters, \eg, the window size and mask sampling times ($T$ and $N_s$ in Eq. \eqref{eq:update_baseline}). The results in Table \ref{tab:hyper-params} demonstrate that being different from with vs. without moving average baseline, small $T$ and $N_s$ can already offer promising performance, further increasing them only produces marginal improvement. In other words, our method is robust to those hyper-parameter values. Considering computational overhead, we choose small $T=5$ and $N_s=2$ throughout our entire experiments.

\begin{table}[h]
\vspace{-1.5mm}
\centering
\small
\setlength{\tabcolsep}{1.5mm}{
\begin{tabular}{@{}l||c||c||c@{}}
\hline
\hline
Method & PruneRate & LLaMA-2-7B & LLaMA-2-13B \\
\hline

w/o MAB & \multirow{2}{*}{30$\%$} & 32.53 & 24.73 \\
with MAB &  & \textbf{28.18} & \textbf{21.99} \\
\hline

w/o MAB & \multirow{2}{*}{40$\%$} & 60.99 & 64.34 \\
with MAB &  & \textbf{39.81} & \textbf{31.52} \\
\hline

w/o MAB & \multirow{2}{*}{50$\%$} & 69.47 & 185.87 \\
with MAB &  & \textbf{65.21} & \textbf{52.23} \\

\hline
\hline
\end{tabular}}
\vspace{-3mm}
\caption{Ablations on the proposed Moving Average Baseline (MAB) in the policy gradient estimator for Channels and heads pruning on LLaMA-2-7B/13B.}
\label{tab:wo_mab}
\vspace{-3mm}
\end{table}

\begin{table}[h!]
\vspace{-2mm}
\centering
\small
\setlength{\tabcolsep}{1.1mm}{
\begin{tabular}{@{}l||c|c|c||c|c|c}
    \hline
    \hline
    \multirow{2}{*}{Hyper-params} &  \multicolumn{3}{c||}{$T$} & \multicolumn{3}{c}{$N_s$} \\
    \cline{2-7}
    &  3 & 5$\star$ & 7 & 2$\star$ & 3 & 4 \\
    \hline
    Perplexity &  21.23 & 21.99 & \textbf{20.08} & 21.99 & 21.71 & \textbf{21.37} \\
    \hline
    \hline
\end{tabular}}
\vspace{-3mm}
\caption{Ablation on the hyperparameters of the moving average baseline, \ie, different window sizes $T$ and mask sampling times $N_s$. Perplexity is tested on the WikiText2 dataset of LLaMA-2-13B with 30$\%$ pruning rate. The hyper-parameter values used in the main results are denoted with $\star$.}
\label{tab:hyper-params}
\vspace{-3mm}
\end{table}

\begin{table*}[h]
\centering
\small

\begin{subtable}{0.5\textwidth} 
  \centering
  \subcaption{Channels and Heads Pruning.}
  \label{tab:channel_proj_strategy}
  \begin{tabular}{@{}l|c|cc@{}}
  \hline
  \hline
    Method & Sparsity & 7B & 13B \\
  \hline
    Sigmoid-Norm  & \multirow{2}{*}{30$\%$} & \textbf{28.18} & \textbf{21.99} \\
    Score-Const &                         & 32.25 & 25.38 \\
    \hline

    Sigmoid-Norm & \multirow{2}{*}{35$\%$} & \textbf{32.52} & \textbf{26.27} \\
    Score-Const &                        & 40.61 & 40.51 \\
    \hline

    Sigmoid-Norm & \multirow{2}{*}{40$\%$} & \textbf{39.81} & \textbf{31.52} \\
    Score-Const &                        & 44.46 & 52.10 \\
    \hline

    Sigmoid-Norm & \multirow{2}{*}{45$\%$} & \textbf{52.07} & \textbf{40.99} \\
    Score-Const &                        & 65.31 & 61.04 \\
    \hline

    Sigmoid-Norm & \multirow{2}{*}{50$\%$} & \textbf{65.21} & \textbf{52.23} \\
    Score-Const &                        & 77.07 & 88.72 \\
  \hline
  \hline
  \end{tabular}
\end{subtable}\hfill 
\begin{subtable}{0.5\textwidth} 
  \centering
  \subcaption{Layer Pruning.}
  \label{tab:layer_proj_strategy}
  \begin{tabular}{@{}l|c|cc@{}}
  \hline
  \hline
    Method & Sparsity & 7B & 13B \\
  \hline
    Sigmoid-Norm  & \multirow{2}{*}{30$\%$} & \textbf{23.20} & 21.93 \\
    Score-Const &                         & 25.32 & \textbf{19.31} \\
    \hline

    Sigmoid-Norm & \multirow{2}{*}{35$\%$} & 33.27 & 26.46 \\
    Score-Const &                        & \textbf{31.37} & \textbf{23.40} \\
    \hline

    Sigmoid-Norm & \multirow{2}{*}{40$\%$} & \textbf{38.26} & 30.99 \\
    Score-Const &                        & 42.30 & \textbf{29.25} \\
    \hline

    Sigmoid-Norm & \multirow{2}{*}{45$\%$} & 69.23 & \textbf{39.26} \\
    Score-Const &                        & \textbf{63.91} & 39.50 \\
    \hline

    Sigmoid-Norm & \multirow{2}{*}{50$\%$} & \textbf{104.37} & 69.92 \\
    Score-Const &                        & 135.51 & \textbf{54.37} \\
  \hline
  \hline
  \end{tabular}
\end{subtable}
\vspace{-3mm}
\caption{Results with \emph{different projection strategies} for pruning heads, channels, and layers on LLaMA-2-7B/13B. Initialization metrics are from Wanda-sp for heads/channels and Layerwise-PPL for layers.}
\label{tab:proj_strategy}
\vspace{-2mm}
\end{table*}

\begin{table*}[h]
\centering
\small
\begin{tabular}{@{}l||c|c||c|c||c|c@{}}
\hline
\hline
Method & PruneRate & Perplexity & PruneRate & Perplexity & PruneRate & Perplexity  \\
\hline

Layerwise-PPL & \multirow{2}{*}{30$\%$} & 24.83 & \multirow{2}{*}{40$\%$} & 41.45 & \multirow{2}{*}{50$\%$} & 126.08 \\
SLEB & & \underline{23.43} & & 40.26 & & 131.49 \\
\hline
\hline
Ours (Random Init) & \multirow{2}{*}{30$\%$} & 26.65 & \multirow{2}{*}{40$\%$} & 42.76 & \multirow{2}{*}{50$\%$} & 125.20 \\
Ours (Random-Prog. Init) & & 30.05 & & \underline{38.28} & & \underline{111.87} \\
\hline
Ours (Layerwise-PPL Init) & 30$\%$ & \textbf{23.20} & 40$\%$ & \textbf{38.26} & 50$\%$ & \textbf{104.37} \\

\hline
\hline
\end{tabular}
\vspace{-3mm}
\caption{Layer pruning results with \emph{different initializations} using LLaMA-2-7B. \textbf{Bold} and \underline{Underscored} denote the first and second best results, respectively.}
\label{tab:layer_diff_init}
\vspace{-6mm}
\end{table*}

\vspace{-3mm}
\subsection{Ablations on Projection Strategy for Initialization: From Metric to Probability} \label{app:proj_strategy}

As the initialization of our \texttt{Bernoulli} policy should be probabilistic values between 0 and 1, but the metrics calculated by the metric-based methods \cite{sun2023simple, an2024fluctuation, ma2023llm} may not hold this range, we thus need to project those metric values to [0, 1] as our initialization. We introduce two projection strategies from metric values $\mathbf{m}$ to probabilities $\mathbf{s}$. The first is called \emph{Sigmoid-Norm} strategy, which is applied in our main experiments:
\begin{equation}
    \mathbf{s} = \text{sigmoid}(\text{Norm}(\mathbf{x}))
\end{equation}

where $\text{Norm}(\cdot)$ is used to linearly normalize the input to a Gaussian distribution with 0 mean and unit variance, then $sigmoid(\cdot)$ is used to transform the input to $[0,1]$. 

An alternative second strategy is named \emph{Score-Const}. It straightforwardly sets mask 1 from metric-based methods as a constant $c$, and mask 0 as $1-c$:
\begin{equation}
s_i = \left\{
\begin{aligned}
& c, & \mathrm{if}\,\, m_i = 1, \\
& 1-c, & \mathrm{if}\,\, m_i = 0,
\end{aligned}
\right.
\end{equation}
The constant $c$ is set to 0.8 in the following experiments, indicating that the initialized \texttt{Bernoulli} probability of the remaining modules is 0.8 and those to be pruned is 0.2. 

The results of different projection strategies on LLaMA-2-7B/13B are detailed in Table \ref{tab:proj_strategy}, which shows that the \emph{Sigmoid-Norm} projection outperforms its \emph{Score-Const} counterpart for most cases. It may be because the order-preserving projection strategy of \emph{Sigmoid-Norm} preserves more information about relative importance among modules, and therefore benefits the optimization.

\vspace{-2mm}
\subsection{More Ablations with Different Initializations} \label{app:diff_init}
\vspace{-1mm}

\textbf{Progressive Pruning with Random (Random-Progressive) Initialization.} Our progressive pruning with random initialization is inspired by the facts that 1) the \emph{continous} \texttt{Bernoulli} probability learned by our method indicates the importance of the corresponding module, therefore the \emph{continous} probability scores from a low pruning rate (\eg, 10\%) encodes fatal information and can be naturally used as the initialization for a higher pruning rate (\eg, 15\%); and 2) the LLMs is likely to exhibit large redundancy when the pruning rate is extremely low (\eg, 5\%), thus random initialization will not significantly degrade the pruning performance (compared to a carefully chosen metric-based pruning initialization) given an extremely low pruning rate such as 5\%. Therefore, to validate our method without a prior metric-based initialization, we propose a progressive pruning strategy, by starting from 5\% pruning rate with random initialization and progressively pruning rate to 50\% by a step size of 5\%. We train this strategy with each pruning rate for 1/3 epoch to maintain efficiency. 

Moreover, Table \ref{tab:layer_diff_init} shows \emph{layer} pruning results with different initializations on LLaMA-2-7B. 

\begin{table*}[ht]
\centering
\small
    \begin{subtable}{0.45\textwidth}
        \centering
        \begin{tabular}{c|l|l@{}}
        \hline
        \hline
            \multirow{2}{*}{\texttt{nsamples}} & \multicolumn{2}{c}{PPL} \\
            \cline{2-3}
            ~ & mean & std \\ \hline
            64 & 27.85 & 1.16 \\ \hline
            128 & 27.94 & 1.54 \\ \hline
            256 & 28.05 & 1.28 \\ \hline
            512 & 28.52 & 1.37 \\ \hline
            1024 & 27.92 & 1.46 \\ \hline
            40k & 27.60 & 1.32 \\ \hline
            120k & 27.19 & 1.18 \\ 
            \hline
            \hline
        \end{tabular}
        \caption{Effect of the number of calibration samples} \label{subtab:abl_nsamples}
    \end{subtable}
    \begin{subtable}{0.45\textwidth}
        \centering
        \begin{tabular}{c|l|l@{}}
        \hline
        \hline
            \multirow{2}{*}{\texttt{seqlen}} & \multicolumn{2}{c}{PPL} \\
            \cline{2-3}
            ~ & mean & std \\ \hline
            128 & 27.94 & 1.54 \\ \hline
            256 & 27.90 & 0.86 \\ \hline
            512 & 28.51 & 0.51 \\ \hline
            1024 & 27.51 & 0.68 \\ \hline
            2048 & 26.68 & 0.64 \\ \hline
            \hline
            \hline
        \end{tabular}
        \caption{Effect of the calibration sequence lengths} \label{subtab:abl_seqlen}
    \end{subtable}
    \vspace{-3mm}
    \caption{Ablations on the number of calibration samples and sequence lengths on PPL (evaluated with 128 sequence length).} \label{tab:abl_nsamples_seqlen}
    \vspace{-3mm}
\end{table*}

\begin{table*}[h]
\centering
\small
\begin{tabular}{@{}l||c||c||c|c|c|c|c||c@{}}
\hline
\hline
Method & PruneRate & PPL $\downarrow$ & PIQA & HellaSwag & WinoGrande & ARC-e & ARC-c & Average \\
\hline

Dense & 0$\%$ & 12.19 & 78.02 & 57.17 & 68.43 & 76.30 & 43.51 & 64.69 \\
\hline

SliceGPT & \multirow{4}{*}{20$\%$} & 24.87 & 74.92 & 49.91 & 66.22 & 69.11 & 35.32 & 59.10 \\
Wanda-sp &  & 23.08 & 77.09 & \textbf{54.34} & 65.90 & 71.21 & \textbf{40.27} & 61.76 \\
Bonsai &  & 23.03 & 76.82 & 53.10 & 64.25 & 71.17 & 39.85 & 61.04 \\
Ours &  & \textbf{19.61} & \textbf{77.09} & 53.45 & \textbf{66.38} & \textbf{72.39} & 40.02 & \textbf{61.87} \\
\hline

SliceGPT & \multirow{4}{*}{30$\%$} & 40.96 & 71.71 & 44.58 & \textbf{64.80} & 60.73 & 30.20 & 54.40 \\
Wanda-sp &  & 42.96 & 74.59 & 48.43 & 59.12 & 63.47 & 34.30 & 55.98 \\
Bonsai &  & 48.30 & 72.85 & 48.25 & 57.77 & 63.8 & 33.87 & 55.31 \\
Ours &  & \textbf{27.13} & \textbf{75.79} & \textbf{49.00} & 62.27 & \textbf{65.36} & \textbf{34.56} & \textbf{57.40} \\

\hline
\hline
\end{tabular}
\vspace{-3mm}
\caption{LLaMA-2-7B pruning results with \emph{the same calibration data} in all methods, evaluated in perplexity (PPL) and accuracies ($\%$) for 5 zero-shot tasks with pruning rates 20\% and 30\%.}
\label{tab:same_calib_data}
\vspace{-5mm}
\end{table*}

\vspace{-2mm}
\subsection{Analysis of the Post-Pruning Modules} \label{app:post-prune}
\begin{figure}[hb]
\vspace{-3mm}
\centering
\begin{minipage}[t]{.5\textwidth}
  \centering
  \includegraphics[width=\linewidth]{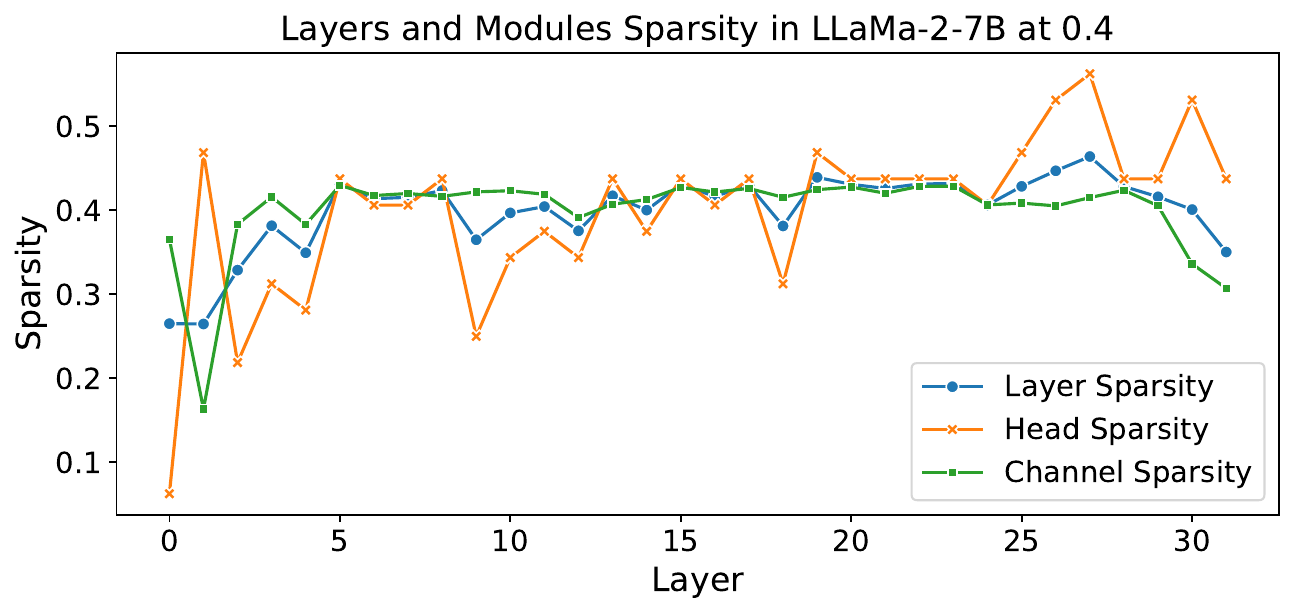}
  \vspace{-8mm}
\end{minipage}

\vspace{-1mm} 
 
\begin{minipage}[t]{.5\textwidth}
  \centering
  \includegraphics[width=\linewidth]{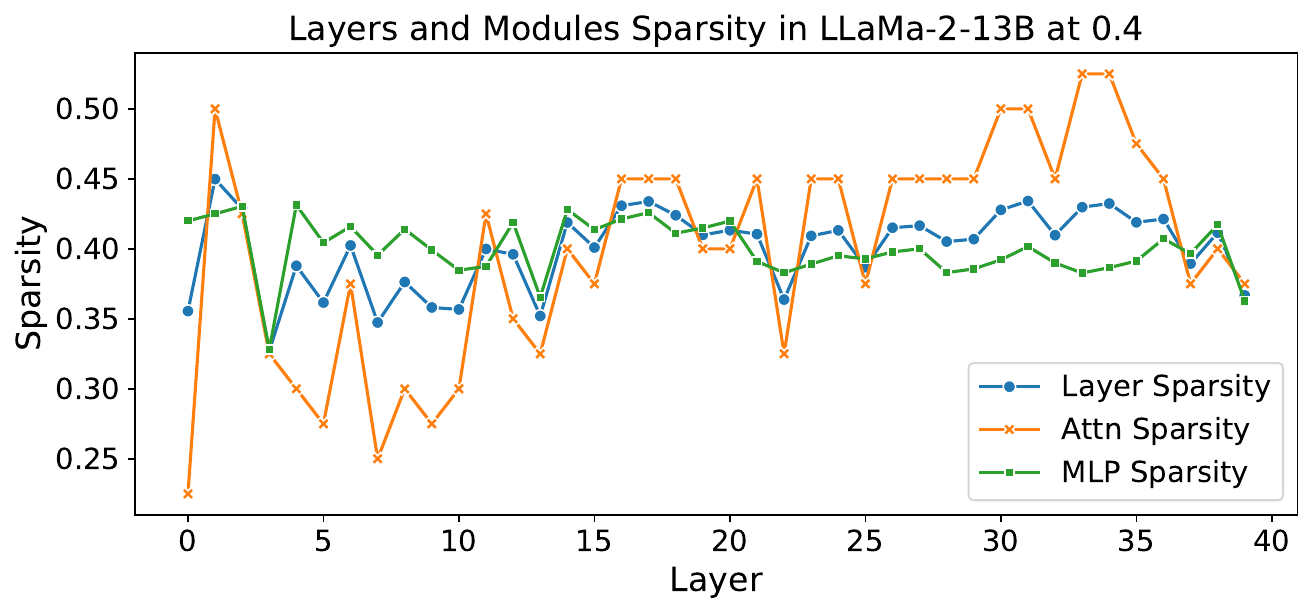}
\end{minipage}
\vspace{-8mm}
\caption{Channels, heads, and layers sparsities of LLaMA-2-\{7B, 13B\} with 40\% pruning rate.}
\label{fig:analysis_sparsity}
\vspace{-5mm}
\end{figure}

\vspace{-1mm}
As global and heterogeneous pruning is performed through our optimization, it is interesting to investigate the pruned modules in each layer. We show the channels, heads, and layers sparsity (\ie, the pruned portion of the corresponding granularity) on LLaMA-2-\{7B, 13B\} with channels and heads pruning at 40\% in Fig. \ref{fig:analysis_sparsity}.

Figures \ref{fig:analysis_sparsity} demonstrate that the pruned LLM exhibits low sparsity in the first and last layers, which is consistent with the previous studies that these layers have a profound impact on the performance of LLMs \cite{ma2023llm}. Moreover, it can be observed that the heads (of MHA) granularity exhibits lower sparsity in the shallow layers (especially in the first layer), while such observation does not hold for the channels (of MLP) granularity. In other words, the pruned sparsity of the channel granularity is more evenly distributed whereas the deeper layers have slightly less sparsity. This might imply that the shallow layers focus more on attention, while the deeper layer imposes slightly more responsibility for lifting the feature dimensions through MLP.

\subsection{Analysis of the Calibration Data Size} \label{app:calib_data_size}
\textbf{Ablations on the number of calibration samples and sequence lengths.} Extensive ablations on the number of calibration samples (\texttt{nsamples}) and sequence lengths (\texttt{seqlen}) are conducted to show the robustness of our method. Specifically, we calibrate our method using C4, and evaluate on WikiText2 on LLaMA2-7B with 30\% sparsity, we report the mean and std over 5 replicates for each experiment. 
\begin{itemize}
\vspace{-3mm}
\item \textbf{Effect of the number of calibration samples} in Table \ref{subtab:abl_nsamples}: we fix \texttt{seqlen} to 128 and vary \texttt{ nsamples} from 64 to 120K.
\vspace{-3mm}
\item \textbf{Effect of the calibration seqence lengths} in Table \ref{subtab:abl_seqlen}: we fix \texttt{nsamples} to 128 and vary \texttt{seqlen} from 128 to 2048.
\vspace{-3mm}
\end{itemize}

The ablations demonstrate that our method is \emph{robust to various calibration data sizes, in terms of both samples\footnote{Reducing \texttt{nsamples} may not save time as our method fixes total training steps (smaller \texttt{nsamples} need more epochs), leading to similar runtime and performance.} and sequence lengths.}

\textbf{Pruning performance using the same calibration data.} We prune LLaMA-2-7B for 20\% and 30\% sparsity with \emph{the same calibration data} across all methods, i.e., \texttt{nsamples} = 128 and \texttt{seqlen} = 2048 from C4. Due to hardware constraints, we exclude LLM-Pruner as it cannot handle sequences of 2048 tokens with backpropagation.

The results demonstrate that \emph{our method consistently outperforms SOTAs using the same calibration data for most cases.}







\begin{table*}[b]
\centering
\small
\begin{tabular}{p{0.1\textwidth}|p{0.85\textwidth}}
\hline
\hline

\multirow{15}{*}{LLM-Pruner} & \underline{Neural Sequential Model, especially transformers,} use a lot of data. We have built a team of experts to help companies reduce the time to develop AI solutions. We can help you from the initial stages in the development of algorithms.
We work in close cooperation with clients to build machine learning and artificial intelligence solutions on a wide spectrum of services, leveraging their own teams or by providing skilled specialists for development and maintenance. \\
\cline{2-2}
 & \underline{AI can create a logo in seconds}, saving business owners the time, cost, and stress of trying to find a logo. If your logo needs a little refreshing or you're looking for a whole brand identity design, Artificial Solutions can help you reach that goal. All projects are completed on time and within budget. Call now!
Clients who want a logo for a family, or non-profit charity, are given a brief description of who they are and what they want and our team of professionals will come up with the logo for you!
A logo design is typically \\
\cline{2-2}
 & \underline{What’s great about the holiday season} is it can be enjoyed with so many of our favorite foods and beverages. I like the idea of coming up with a drink that is special for the holidays and toasts those celebrating and visiting. The holiday cocktail is a great time to express your individuality and get friends together in one place to mingle and enjoy delicious cocktails.
I asked the cocktail geniuses over at Flaver Liquor Bar what their favorite cocktail is and one that got me excited was The Pink Lemon \\
\hline
\hline

\multirow{16}{*}{SliceGPT} & \underline{Neural Sequential Model, especially transformers,} provides one of the best ways of creating an attractive, clean, and appealing image of your body. Using 2.0 mHz and 20mhZ transomters on the face can be very effective.
Here we use the VAS 2.0, 20 mhc, and VAS 20 with Slim for extra definition and volume.
The model is a 19mm tube that is about the size of an 5mm tube.
The tube is 1.72mm  \\
\cline{2-2}
 & \underline{AI can create a logo in seconds} and a logo is a valuable marketing asset for your business.
In general, all the digital devices are not a good option for those who are a bit low on budget. A basic website design and the creation of a logo can be affordable for the business. Most of the businesses offer the services, especially to the small companies.
The website design is a significant matter to develop a business image for the companies. A logo or logo design helps in this process to create the image. There is a good number of companies and designers who can build a logo within hours. \\
\cline{2-2}
 & \underline{What’s great about the holiday season}? Well, there are many things: the holidays in the U.S are celebrated in many ways, which is wonderful for the family. Besides the festivities, there are also plenty of goodies and delicious foods to enjoy. We can all enjoy our favorite foods in celebration of Christmas. What do we really enjoy the most during the holidays? There are definitely more interesting things to celebrate in the season of holidays, but we also look forward to the holiday time because the holiday means the delicious food that you can enjoy. \\
\hline
\hline

\multirow{17}{*}{Ours} & \underline{Neural Sequential Model, especially transformers,} have been shown to be efficient in time and space complexity as well as delivering great recognition results on a single machine. These are used frequently in many domains of computer vision, including human body detection, motion capture and autonomous driving. Although they achieve relatively high performance, their performance is bounded by the dimension space they can operate on. In this paper, we describe the novel concept of a compact transformer based on a neural sequence that combines the advantages of transformers and neural networks. Our analysis shows that the compact transformer can process the images in their entirety while inc  \\
\cline{2-2}
 & \underline{AI can create a logo in seconds} – at a lower cost than ever. What’s the problem with this? Well, you don’t get that perfect custom logo you have always wanted. Not the way it will work in your business, anyway.
In-house logo creation, however, can be a bit time-consuming. You’ll need a designer to help you out. You will also need to know the font and design you like most. You may have some logo ideas in your head, but without tools and expertise at your fingertips, you’ll have to work hard for them \\
\cline{2-2}
 & \underline{What’s great about the holiday season} is that it’s a time for us all to relax and spend some time with our friends and families. At this time of year, we all like to share stories that bring us closer together. One way to do this is through gifts. Here are a few suggestions that we hope you’ll take with you into the new year.
You can give your best loved something in the form of memories or you can get them a gift that will have lasting impact, like this new watch.
We’re thrilled to introduce the new Seiko 6, an \\
\hline
\hline

\end{tabular}
\vspace{-3mm}
\caption{Generated samples of the pruned LLaMA-2-7B model with 30\% pruning rate by different methods.}
\label{tab:gen_samples}
\end{table*}

\end{document}